\documentclass[acmsmall]{acmart}
\AtBeginDocument{%
  }

\usepackage{cc,lambda}
\usepackage[noshare]{vlc}
\usepackage{proof}
\usepackage{pdfpages}
\usepackage{multirow}
\usepackage{xargs}

\usepackage{mathpartir}
\usepackage{mathtools}
\usepackage{xcolor}
\usepackage{calc}
\usepackage[labelformat=simple]{subcaption}

\usepackage{paralist}
\usepackage{tikz}
\usepackage{booktabs}
\usepackage{pifont}
\usepackage[inline]{enumitem}
\usepackage[utf8]{inputenc}
\usepackage[english]{babel}
\usepackage{scalerel,amsmath}
\usepackage{stmaryrd}
\usepackage{wrapfig}

\usepackage{listings}

\usepackage{amsthm,amssymb}
\usepackage{listingsutf8} 
\newcommand{\err}{\OB{\bot}}

\newcommand{\rev}[1]{\setlength{\dummylen}{\fboxrule}\setlength{\fboxrule}{2pt}%
            \vspace{1ex}\noindent\hfill%
            \fbox{\begin{minipage}{.96\columnwidth}Added by Sheng: \\ #1\end{minipage}}%
            \setlength{\fboxrule}{\dummylen}\hfill{}\vspace{1ex}}

\tikzset{
	dot/.style = {circle, minimum size=#1,
		inner sep=0pt, outer sep=0pt},
	dot/.default = 10pt % size of the circle diameter 
}

\newcommand{\parag}[1]{\smallskip\noindent\textbf{#1}\ \ }

\newcommand{\tp}{\ensuremath{\pi}}

\renewcommand{\to}{\ensuremath{\operatorname{\rightarrow}}}

\newcommand{\TypeEnvF}{\ensuremath{\ddot{\TypeEnv}}}

\newcommandx{\hasTypeVGC}[4][1=\tp,2=\TypeEnv]{\OB{#2\vdash_C\ofType{#3}{#4}}}

\newcommand{\prog}[1]{{\progfontsize\texttt{#1}}}

\newcommand{\vstaticer}{\ensuremath{\Omega}}

\newcommand{\configSym}{\ensuremath{K}}
\newcommand{\config}{\configSym}

\newcommandx{\hasTypeG}[3][1=\TypeEnv]{\OB{#1\vdash_{\textit{G}}\ofType{#2}{#3}}}
\newcommandx{\hasTypeUpdate}[5][1=\tp,2=\config,3=\TypeEnv]{\OB{#1;#2;#3\vdash_{\textit{U}}\ofType{#4}{#5}}}
\newcommandx{\castInsert}[4][1=\TypeEnvF]{\OB{#1\vdash_{\textit{I}}\ofType{#2 \leadsto #3}{#4}}}
\newcommandx{\castInsertM}[4][1=\TypeEnvF]{\OB{#1\vdash_{\textit{M}}\ofType{#2 \leadsto #3}{#4}}}
\newcommandx{\castInsertS}[4][1=\TypeEnv]{\OB{#1\vdash_{\textit{S}}\ofType{#2 \leadsto #3}{#4}}}

\newcommandx{\hasTypeVG}[5][1=\vstaticer,2=\TypeEnv,3=\tp]{\OB{#3;#2\vdash\ofType{#4}{#5}\,|\, #1}}
\newcommandx{\hasTypeErr}[4][1=\tp,2=\TypeEnv]{\OB{#1;#2 \vdash \ofType{#3}{#4}}}

\newcommand{\tycompSym}{\ensuremath{\approx}}

\newcommandx{\tycompc}[3][1=\tp]{\ensuremath{#2 \tycompSym_{#1}^? #3}}

\newcommand{\dummyloc}{\ensuremath{\ell}}

\newcommand{\ploc}[1]{\ensuremath{\dummyloc_{\prog{#1}}}}

\newcommand{\preciseSym}{\ensuremath{\sqsubseteq}}
\newcommand{\precise}[2]{\ensuremath{#1 \preciseSym #2}}

\newcommand{\NOTESolved}[1]{}
\newcommandx{\realCast}[4][1={}]{\ensuremath{#2:#3 \Rightarrow^{#1} #4  }}
\newcommandx{\realCastRgt}[3][1={}]{\ensuremath{:#2 \Rightarrow^{#1} #3  }}
\newcommandx{\prealCast}[4][1={}]{\ensuremath{\prog{#2}:#3 \Rightarrow^{\ploc{#1}} #4  }}
\newcommandx{\prealCastRgt}[3][1={}]{\ensuremath{:#2 \Rightarrow^{\ploc{#1}} #3  }}

\newcommandx{\threeCast}[6][1={},2={}]{\ensuremath{#3:#4 \Rightarrow^{#1} #5  \Rightarrow^{#2} #6}}

\newcommandx{\allFun}[3][1=\alpha,2=\cst]{\ensuremath{\forall #1.\fun{#2}{#3}}}

\newcommandx{\expPrecision}[4][1=\TypeEnv, 2=\TypeEnv']{\ensuremath{#1;#2 \vdash \precise{#3}{#4}}}

\def\draft{1}

\if\draft1

\else

\fi

\newcommand{\UNSAT}{\textbf{UNSAT}}
\newcommand{\SATLabel}{\textbf{SAT}}
\newcommand{\SATName}{SAT}

\usepackage{listings}
\usepackage[T1]{fontenc}
\usepackage{minted}

\newcounter{experiment}

\definecolor{codegreen}{rgb}{0,0.6,0}
\definecolor{codegray}{rgb}{0.5,0.5,0.5}
\definecolor{codepurple}{rgb}{0.58,0,0.82}
\definecolor{backcolour}{rgb}{0.95,0.95,0.92}
\lstdefinestyle{promptstyle}{
    backgroundcolor=\color{backcolour},
    basicstyle=\ttfamily\footnotesize,
    columns=fullflexible,
    breaklines=true,
    breakatwhitespace=false,
    captionpos=b,
    keepspaces=true,
    numbers=left,
    numberstyle=\scriptsize\color{codegray},
    numbersep=4pt,
    frame=single,
    framesep=2pt,
    xleftmargin=2em,
    xrightmargin=2em,
    showspaces=false,
    showstringspaces=false,
    showtabs=false,
    tabsize=2
}

\usepackage[dvipsnames]{xcolor}
\usepackage[normalem]{ulem}

\renewcommand{\rev}[1]{\textcolor{BrickRed}{#1}}

\newcommand{\revdel}[1]{\textcolor{Gray}{\sout{#1}}}
\renewcommand{\rev}[1]{#1}
\renewcommand{\revdel}[1]{}

\setcopyright{cc}
\setcctype{by}
\acmDOI{10.1145/3808209}
\acmYear{2026}
\acmJournal{PACMSE}
\acmVolume{3}
\acmNumber{FSE}
\acmArticle{FSE202}
\acmMonth{7}
\acmSubmissionID{fse26mainb-p2731-p}
\received{2025-09-04}
\received[accepted]{2026-03-24}

\begin{document}

% \title{Evaluating Satisfiability Solving with LLMs}
\title{Satisfiability Solving with LLMs}
\subtitle{A Matched-Pair Evaluation of Reasoning Capability}

\begin{abstract}
Large language models (LLMs) are increasingly used for tasks that implicitly reduce to Boolean satisfiability (SAT), yet their \emph{reasoning} ability on SAT remains unclear. We present a systematic study of LLMs on 2-SAT and 3-SAT, together with two canonical reductions—Vertex Cover and a discrete 3D-packing formulation—designed to probe \emph{representation-invariant} reasoning. Our evaluation begins with the conventional lens (accuracy/precision/recall/F1) and the phase-transition setting. We find that traditional metrics are frequently misleading. Models achieve high scores even \revdel{through} \rev{though} they tend to classify \revdel{al} \rev{all} formulas as satisfiable, fail to reproduce
the classical easy--hard--easy signature around the 3-SAT threshold, and degrade sharply as the number of
variables $N$ grows.

To address this, we introduce a \emph{paired-formula} protocol (minimally different satisfiable/unsatisfiable instances) and a new measure, \emph{Accurate Differentiation Rate} (ADR), which requires prediction on \emph{both} members of each pair correct. ADR cleanly separates reasoning-oriented models from heuristic ones and correlates with \emph{witness validity} (truth assignments that actually satisfy the formula). Extending beyond CNF, we test cross-representation consistency via standard reductions: (i) Convert CNF to Vertex Cover and (ii) Convert 3-SAT to discrete 3D packing with verifiable placement constraints. Decisions made on CNF and on their graph/packing counterparts agree for most models on $>\!80\%$ of instances, revealing stable decision rules across representations. A leading model (e.g., GPT-5) achieves both high invariance and correctness on small $N$, but still suffers scale-induced degradation.

Taken together, our results support the thesis that SAT is a 
\revdel{foundational}\rev{\emph{conservative}} probe for LLM reasoning: performance on SAT predicts transfer to other NP-style reductions, while paired evaluation with ADR provides a faithful, representation-robust assessment beyond conventional metrics.
% \rev{Taken together, our results position SAT as a \emph{conservative probe} for LLM reasoning: under two canonical NP-complete reductions (Vertex Cover and discrete 3D packing), models largely preserve their SAT-era behavior and do not show systematic gains over the SAT baseline. ADR remains a faithful, representation-robust assessment when conventional metrics are confounded by imbalance.}

\end{abstract}




\author{Leizhen Zhang}
\orcid{0009-0007-4141-1155}
\affiliation{%
  \institution{University of Louisiana at Lafayette}
  \city{Lafayette}
  \country{USA}
}
\email{leizhen.zhang1@louisiana.edu}

\author{Shuhan Chen}
\orcid{0009-0002-5869-3949}
\affiliation{%
  \institution{East China Normal University}
  \city{Shanghai}
  \country{China}
}
\email{10225102449@stu.ecnu.edu.cn}

\author{Sheng Chen}
\orcid{0000-0003-1735-0704}
\affiliation{%
  \institution{University of Louisiana at Lafayette}
  \city{Lafayette}
  \country{USA}
}
\email{sheng.chen@louisiana.edu}

\renewcommand{\shortauthors}{Zhang, Chen, and Chen}


\maketitle

\section{Introduction}
\label{sec:Intro}

Thanks to network architecture and attention mechanism~\cite{Attention:Vaswani:2017,ETA:Tay:2022},
large language models (LLMs) possess strong capabilities to understand
contextual relationships among elements within a sequence~\cite{What:Clark:2019}. Such capabilities
enable LLMs to understand not only syntax, but also semantics and context of
relevant knowledge, enabling them to excel in solving a wide
range of problems. 

Reasoning capability is crucial for solving any nontrivial problems. 
This paper investigates LLMs' capabilities of solving Boolean satisfiability
problems (SAT), both 2SAT and 3SAT.
% where the former can be solved in polynomial time and the latter is NP-Complete. 
%
At first sight, it may seem counterintuitive to study this ability,
% to solve SAT formulas with LLMs, 
\revdel{since there are many efficient algorithms for 2SAT~\cite{Aspvall:1979,Tarjan:1972} and 3SAT~\cite{Davis:1962,Marques:2002,Moskewicz:2001,Een:2003,BMC:2009:Biere}. }
\rev{since 2SAT can be solved in polynomial time~\cite{Aspvall:1979,Tarjan:1972} and modern SAT solvers can handle many large 3SAT instances efficiently in practice~\cite{Davis:1962,Marques:2002,Moskewicz:2001,Een:2003,BMC:2009:Biere}.}
However, for a few reasons, knowing this ability is useful.

\revdel{First, LLMs are more convenient to use than SAT solvers.}
\rev{First, LLMs are often more convenient to use than standalone SAT solvers.}
% \rev{First, LLMs lower the barrier to \emph{attempting} SAT solvers.}
LLMs are easily accessible, 
\rev{whereas SAT solvers require inputs in a formal constraint language (such as CNF/DIMACS).}
\revdel{but each algorithmic SAT solver places specific format requirements on its inputs.} 
To be solved by SAT solvers, relevant problems need to be first translated into the required format before they can be solved. This involves introducing boolean literals to represent objects in the problem and expressing
relations between objects with connectives in SAT formulas. 
\revdel{This translation step is often quite involved (like parsing~\cite{CPTT:2007:Alfred} in compilation) and can be very tricky. }
\rev{Although translators and modeling frameworks exist for certain domains, using them still requires selecting appropriate abstractions and bounds, validating that the encoding captures the intended semantics, and debugging subtle modeling errors, a process that is similar to building and validating a front-end~\cite{CPTT:2007:Alfred} in a compiler toolchain.}
\revdel{As a result, LLMs are essentially available to anyone who has an Internet connection, while SAT solvers are restricted to domain experts.}
\rev{As a result, while SAT solvers are widely available and highly efficient once properly instantiated, their effective deployment often demands domain knowledge and engineering effort. In contrast, LLMs lower the barrier by allowing users to describe constraints and obtain candidate solutions directly in natural language. }
% \rev{As a result, while solver pipelines are mature, using them end-to-end still typically requires nontrivial modeling and validation effort; we therefore use SAT solving by LLMs as a \emph{capability probe}, not as a replacement for solvers.}

% \rev{First, LLMs lower the barrier to \emph{attempting} SAT-style reasoning: they are widely accessible and accept
% natural-language (or lightly structured) problem descriptions. In contrast, solver pipelines typically require
% problem-specific encodings (e.g., CNF/SMT) and careful validation of those encodings. While mature modeling and
% translation tools exist, producing correct encodings and interpreting solver results still demands nontrivial
% expertise. We therefore study end-to-end SAT solving by LLMs as a \emph{capability probe}---not as a replacement
% for modern solvers---and we discuss solver-integrated, neuro-symbolic workflows as future work.}

\revdel{In fact, LLMs have already been widely used to solve many problems[38, 50], 
and many such problems are SAT in disguise.}
% \rev{LLMs have been widely used to \emph{attempt} many combinatorial problems~\cite{IALL:2025:Raza,KACP:2025:Zheng}, and many such problems admit polynomial-time reductions to SAT (or employ SAT solving as a subroutine).}
\rev{In fact, LLMs have already been widely used to solve many constraint-heavy problems~\cite{IALL:2025:Raza,KACP:2025:Zheng}, and a substantial subset of these tasks can be naturally framed as satisfiability or optimization under constraints.}
For example, scheduling with
ternary constraints (such as at most two tasks of X, Y, and Z may be run simultaneously)~\cite{Even:1975}, 3D packing (such as Object X must be next to Y or Z)~\cite{fekete2004combinatorial}, robotic movement planning (such as avoiding collisions between Objects X, Y, and Z)~\cite{Kautz:1996}, 
network configurations~\cite{beckett2017general}, and
Sudoku puzzles~\cite{Yato:2003} can all be encoded as \rev{and are not easier than} 3SAT.
% \rev{can be encoded as SAT/3SAT}~\cite{Cook:2023} problems. 
Simplified versions where constraints are pairwise can be encoded in 2SAT, such as employee shift scheduling and classroom/course timetabling.

\revdel{As a result, LLMs' performance in SAT solving is a strong indication about
their performance in solving such problems.}
% \rev{As a result, SAT solving provides a stringent, representation-controlled \emph{baseline} for assessing whether an LLM exhibits solver-like logical competence that could plausibly transfer to SAT-reducible tasks.}

Second, LLMs are easier to leverage for developing complex analyses.
% many problems that are being solved by LLMs are hard to develop
% exact analyses for. 
Some of such analyses include a step that uses
SAT solving or is equivalent to SAT solving. For example, 
\revdel{accurate}
\rev{semantics-faithful} vulnerability detection needs to reason about
data-flow analysis~\cite{Reps:1995:PID}, alias analysis~\cite{UA:1994:Ramalingam},
path feasibility~\cite{BMC:2009:Biere},
and constant propagation~\cite{Wegman:1991:CPC}.
\revdel{Each is known to be NP-complete, equivalent to SAT solving.}
\rev{These analyses are computationally hard due to feasibility conditions 
involved. Practical tools often 
%often introduce abstractions and bounds and 
discharge such conditions using SAT/SMT backends with potentially iterative refinement loops.}
% Each is known to be NP-complete or harder, equivalent to SAT solving.
%
Numerous efforts~\cite{Ball:2002:SLAM,Henzinger:2002:LazyAbs,Engler:2001:Bugs,shankar2001detecting,Xie:2006:Buffer,Bessey:2010:Coverity,Calcagno:2015:Infer} have been made from the static analysis
perspective, but 
% the machine learning and LLM-based approaches~\cite{}
% reported to perform much better. 
the interest in detecting
vulnerability using LLMs has gained tremendous momentum,
illustrating the shift from solving problems using more
rigorous, rule-based analyses to more opportunistic, data-based analyses.
This shift largely reflects the simplicity of developing analyses with LLMs.
This shift is also happening in other areas, such as program repairing, 
invariant generation and reasoning~\cite{CCSS:2024,RLLI:2005,CLLM:2023,LGIB:Pirzada:2024}, program verification~\cite{EPSS:2024:Wen,Lemur:2023:Wu}, and many others~\cite{CSL:Wang:2025,LLM:Hou:2024}. 

\revdel{For such analyses, the performance of LLMs is unlikely to transcend that on SAT solving.}
% \rev{For such analyses, SAT-level performance offers a conservative diagnostic: if an LLM fails on the underlying constraint reasoning, downstream success is unlikely to be reliable without additional structure, tooling, or verification. We therefore empirically test (rather than assume) how SAT-era behavior carries over under standard reductions.}
% \rev{In these settings, SAT competence is a natural limiting factor whenever the analysis pipeline relies on satisfiability checks as a core subroutine.}

Third, while LLMs and MLs are commonly used in analyses,
evaluating performance of such analyses is not easy, 
due to reasons such as dataset, model parameters, 
randomness. and class imbalance. For example, in vulnerability detection 
previous work reported remarkable performance ~\cite{VulDeePecker:Li:2018, Devign:Zhou:2019, zou2019mu, DeepWukong:Cheng:2021}
but more recent work revealed that such performance
is largely due to noise in existing datasets~\cite{VDCL:Ding:2024}.
For instance, the F1 of StarCode2~\cite{SC2:Lozhkov:2024} 
decreased from 0.68 for a previous dataset to 0.03 for a more accurate dataset and the performance of GPT-3.5 and GPT-4 is similar to random guessing, even with advanced training techniques~\cite{VDCL:Ding:2024}. 

Given the versatility of SAT and that many problems are directly equivalent to or employ SAT as a building block, understanding the ability of SAT solving with LLMs is critical.
\revdel{In particular, their performance on SAT solving can serve as a guidance of that on other problems.}
\rev{In particular, we hypothesize that LLM performance on SAT provides a conservative upper bound for performance on a broad class of constraint-based tasks, based on the following rationale.}

\rev{Consider a constraint-based pipeline that works by (i) deterministically translating a problem instance into one or more SAT formulas, (ii) deciding satisfiability/unsatisfiability of the formula(s), and (iii) deterministically mapping the SAT result back to a solution of the original problem. The pipeline’s end-to-end success rate is upper-bounded by the success rate on the required SAT instances. A correct answer is possible only if the SAT subproblems are answered correctly and additional steps such as encoding/decoding and iterative refinement are all free of failure. }

\rev{Moreover, to generalize robustly in practical settings, ML models must avoid relying on superficial cues and instead reason about the underlying relations and constraints. For many NP-hard/complete tasks, this entails solving an underlying constraint-satisfaction core that can often be encoded as SAT formulas.
For example, for vulnerability detection in practical settings, a model has to reason about various data- and control-flows and pointer and alias relations (NP-hard), rather than the presence of certain macros and conditionals (Please see the performance discrepancies discussed a few paragraphs above). 
As another example, to solve vertex cover, a model has to reason about 
which vertices cover all edges (NP-hard), rather than relying on
statistics such as density or degree distributions.
}

\rev{Thus, when LLMs are used for such tasks in a semantics-faithful manner, their role increasingly resembles the constraint-based pipeline described above. The performance relationship between the SAT decisions in step (ii) and the pipeline’s end-to-end success also applies to LLM-based solutions.
In addition to this informal argument, we empirically test the hypothesis in the later sections of the paper.
}

This paper makes the following contributions. 

\begin{enumerate}
\item We study the problem of SAT solving with LLMs (Sections~\ref{sec:misapprehension} and~\ref{sec:unsat-detection}). Our study 
involves 12 models, including more recent ones such as GPT-5, and 
more than 2700 3SAT formulas with varying 
numbers of variables and difficulties. Our evaluation indicates that
current large language models still have very limited abilities of
solving satisfiability problems, particularly when the number of 
variables is more than 25. 

\item We observe the shortcomings with traditional metrics
for measuring machine learning tools, including LLMs, and propose
the idea of using paired formulas and a new metric named 
ADR (\revdel{Accurate Differentiate Ratio}\rev{Accurate Differentiation Rate}) to address the problem (Section~\ref{sec:paired-formulas}). 
Unlike conventional accuracy or F1, which can be inflated by dataset imbalance and superficial cues, ADR directly evaluates whether a model can consistently distinguish paired satisfiable/unsatisfiable instances, thus offering a reasoning-oriented measure. 
The experiments reveal that ADR precisely captures the LLMs'
abilities of satisfiability solving.

\item To test the usefulness of ADR, we also investigate 
an easier logical reasoning problem, namely 2SAT (Section~\ref{sec:twosat}). Experiments 
with more than 700 generated 2SAT formulas 
show that ADR precisely reflects LLM's abilities of solving this
category of logical reasoning problems. Specifically, many 
reasoning based models, such as GPT 5, DeepSeek-reasoner, and o1, 
achieve a very high ADR even for problems with more than 50 variables. 

\item \revdel{To test our thesis that LLMs' performance on SAT solving 
is an upper bound of that on problems that are equivalent to or use SAT solving as a building block, we investigate the performance of using LLMs for solving two problems: vertex cover (Section~\ref{sec:vc}) and 3D-packing (Section~\ref{sec:packing}). Our evaluation results substantiate this thesis.} \rev{To test the hypothesis that SAT acts as a conservative probe for SAT-reducible tasks, we evaluate the same
paired instances with two canonical reductions from SAT to Vertex Cover (Sec.~\ref{sec:vc}) and discrete 3D packing (Sec.~\ref{sec:packing}). We find that LLMs' prediction results for SAT instances and the corresponding counterparts in vertex cover and 3D packing are highly consistent, often greater than $80\%$. For those the predictions differ, the results for SAT are more likely to be correct than for VC instances and are equally likely to be correct for 3D packing instances. In both tests, ADR remains the most diagnostic
signal across representations.}

\end{enumerate}

The rest of the paper is organized as follows. We present 
the necessary background in Section~\ref{sec:background}, 
discuss related work in Section~\ref{sec:rw},
discuss threats to validity in Section~\ref{sec:threats},  and conclude in Section~\ref{sec:con}.

\section{Background}

\label{sec:background}

\subsection{SAT: Propositional Logic and Conjunctive Normal Form}
\paragraph{Propositional syntax.}
Let $\mathcal{X}=\{x_1,\ldots,x_n\}$ be a finite set of propositional variables.
A \emph{literal} is either a variable $x$ (a positive literal) or its negation $\neg x$ (a negative literal).
A \emph{clause} is a finite disjunction of literals, e.g., $(\ell_1 \lor \ell_2 \lor \cdots \lor \ell_k)$.
A propositional \emph{formula $\varphi$ can always be converted into a conjunctive normal form (CNF)}, which is a finite conjunction of clauses:
\[
\varphi \;=\; \bigwedge_{i=1}^{m} C_i
\qquad\text{with}\qquad
C_i \;=\; \bigvee_{j=1}^{k_i} \ell_{ij}.
\]
An \emph{assignment} $\sigma : \mathcal{X}\to\{\mathsf{false},\mathsf{true}\}$ 
assigns $\mathsf{false}$ or $\mathsf{true}$ to each variable. It
extends to literals and clauses in the usual way.
A formula $\varphi$ is \emph{satisfiable (\SATLabel)} if there exists 
a $\sigma$ such that assigning variables in $\varphi$ according to 
$\sigma$ makes $\varphi$ evaluate to $\mathsf{true}$. 
Otherwise $\varphi$ is \emph{unsatisfiable (\UNSAT)}.\footnote{We write \SATLabel\ (\UNSAT) to mean a formula is satisfiable (unsatisfiable) and write \SATName\ for all other contexts.}

\paragraph{$K$-CNF and problem classes.}
If every clause has exactly $K$ literals ($k_i=K$ for all $i$), the formula is a \emph{$K$-CNF}.
The decision problems \textsc{2SAT} and \textsc{3SAT} ask whether a given $2$-CNF or $3$-CNF is satisfiable.
It is classical traditional that \textsc{2SAT} is solvable in polynomial time (via implication graphs and strongly connected components)~\citep{Aspvall:1979}, 
whereas \textsc{3SAT} is NP-complete~\citep{Cook:2023}.
Throughout, we write $N=|\mathcal{X}|$ for the number of variables and $L$ for the number of clauses.

\subsection{SAT Solvers in Brief}
Modern complete SAT solvers descend from the DPLL framework~\citep{Davis:1962} and its Conflict-Driven Clause Learning refinement (CDCL)~\citep{Marques:2002}.
At a high level they interleave:
(i) \emph{decision} steps that assign a value to an unassigned variable;
(ii) \emph{unit propagation} that deterministically forces literals when a clause becomes unit;
(iii) \emph{conflict analysis} to derive a \emph{learned clause} (often using the 1-UIP scheme) and perform non-chronological backtracking; and
(iv) \emph{restarts} and \emph{branching heuristics} (e.g., VSIDS-style activity scores).
Efficient data structures such as \emph{two-watched literals} make propagation nearly linear in the number of clause occurrences.
Pre- and in-processing (e.g., subsumption, self-subsuming resolution, blocked-clause elimination, bounded variable elimination) further shrink instances.
For \textsc{2SAT}, linear-time algorithms operate on the implication graph: $\varphi$ is SAT iff no variable and its negation are in the same strongly connected component~\citep{Aspvall:1979}.
In contrast, \textsc{3SAT} generally requires the full CDCL arsenal and exhibits the well-known “easy–hard–easy” pattern described next.

% \begin{figure}[t]
%   \centering
%   \includegraphics[width=0.8\linewidth]{figures/figures_CDCL_phase_transition/Random_3-SAT_CDCL_N_75_median.pdf}
%   \caption{Phase transition for random $3$SAT with $N{=}75$ using a CDCL solver. 
%   As the clause-to-variable ratio $\alpha = L/N$ increases, the median number of branching decisions (black, left axis) follows an easy--hard--easy pattern, 
%   reaching its maximum near the critical threshold $\alpha_c \approx 4.26$. 
%   In contrast, the probability of satisfiability (blue, right axis) remains close to $1$ for small $\alpha$, then undergoes a sharp decline around $\alpha_c$, 
%   and eventually approaches $0$ as $\alpha$ grows larger.}
%   \label{fig:phase-cdcl-n200}
% \end{figure}
% Code/invoke_traditional_methond/phase_transition_generate_and_draw_Minisat22_only_draw_median.py

\begin{figure*}[t]
  \centering
  % 左边图
  \begin{subfigure}[t]{0.48\linewidth}
    \centering
    \includegraphics[width=\linewidth]{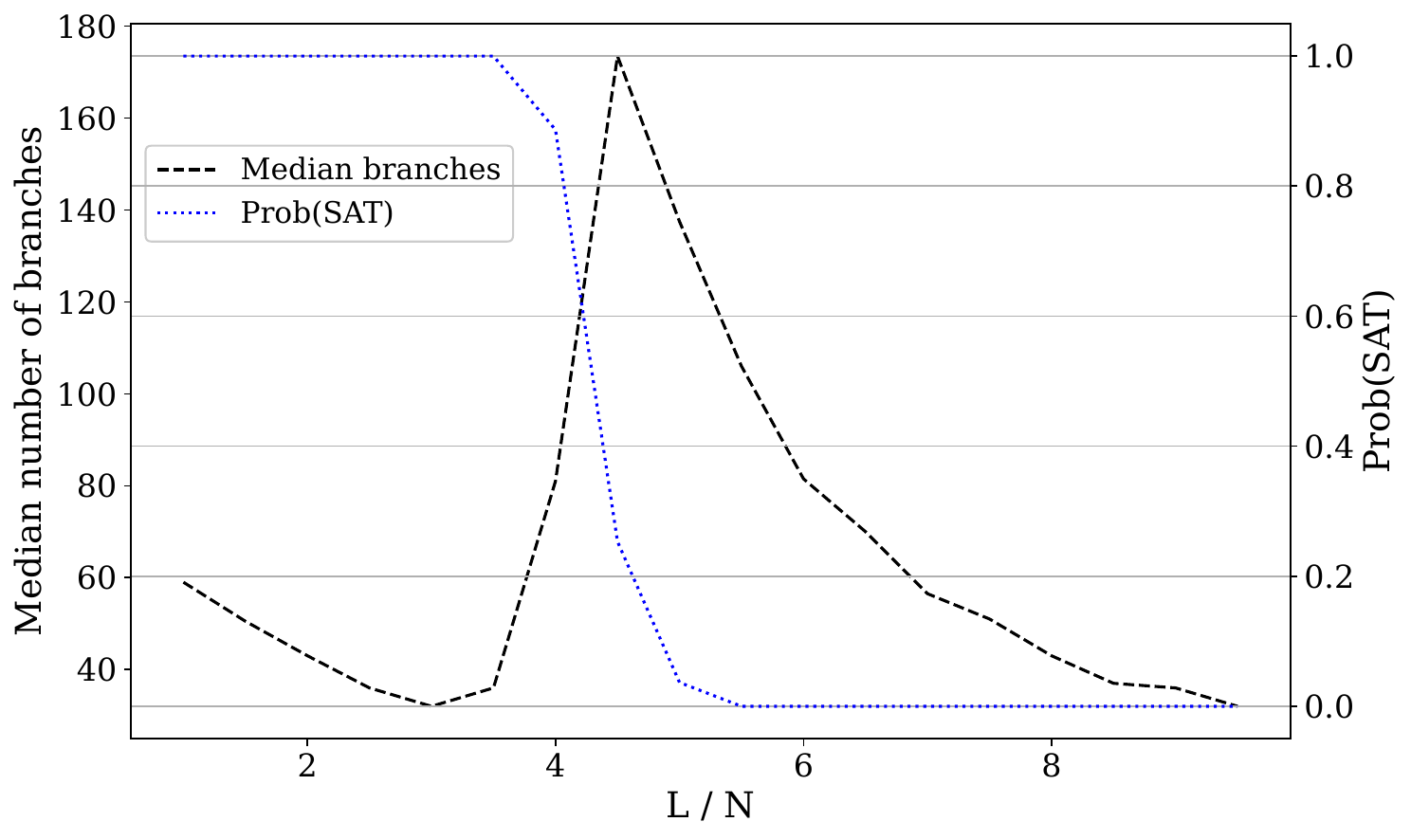}
    \caption{Phase transition for random $3$SAT with $N{=}75$ using a CDCL solver. 
    As the clause-to-variable ratio $\alpha = L/N$ increases, the median number of branching decisions (black, left axis) follows an easy--hard--easy pattern, 
    reaching its maximum near the critical threshold $\alpha_c \approx 4.26$. 
    In contrast, the probability of satisfiability (blue, right axis) remains close to $1$ for small $\alpha$, then undergoes a sharp decline around $\alpha_c$, 
    and eventually approaches $0$ as $\alpha$ grows larger.}
    \label{fig:phase-cdcl-n200}
  \end{subfigure}
  \hfill
  % 右边图
  \begin{subfigure}[t]{0.50\linewidth}
    \centering
    \includegraphics[width=\linewidth]{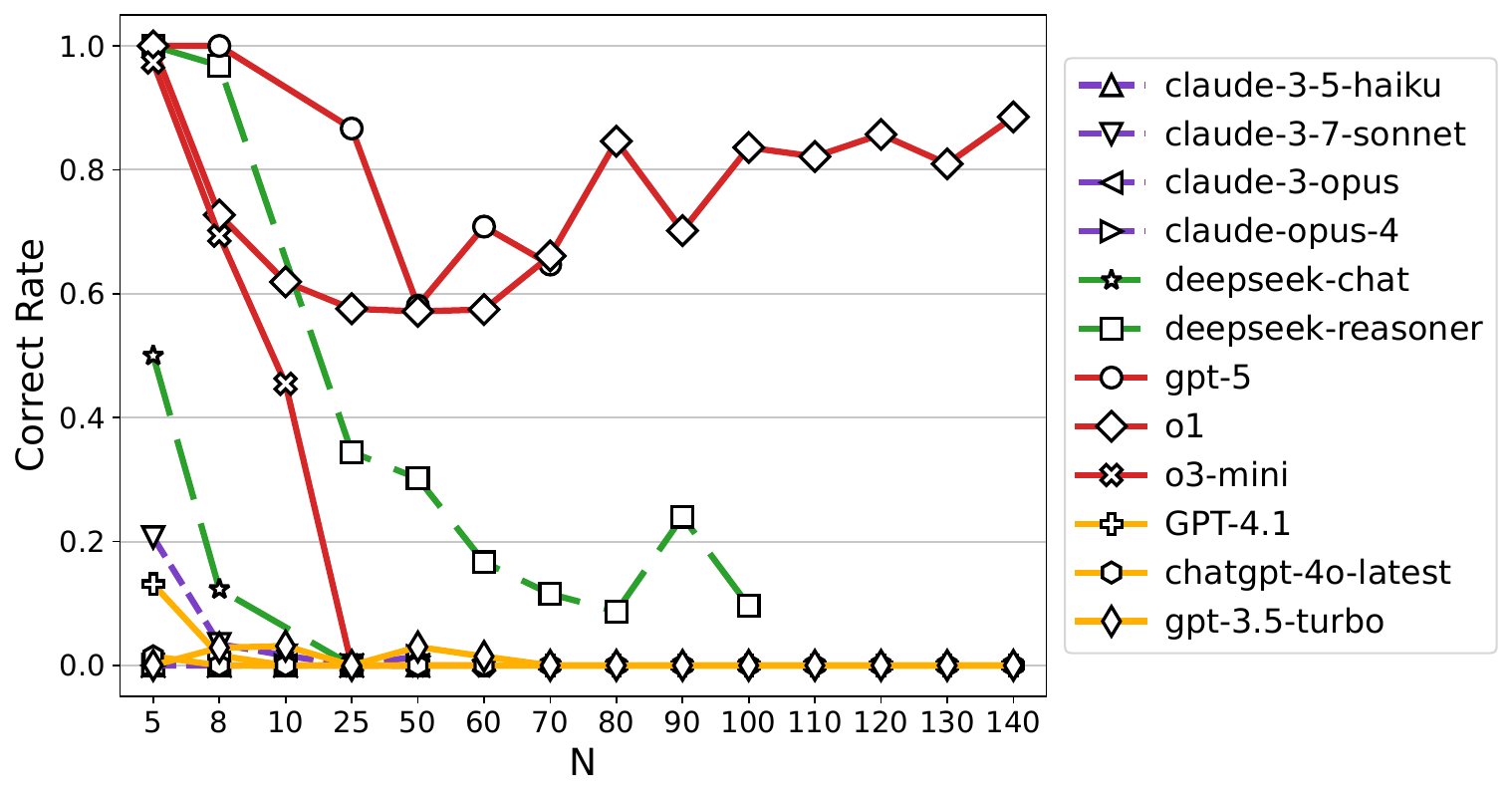}
    \caption{Correct rate of low-$\alpha$ \UNSAT\ CNFs’ prediction across models as $N$ increases. Compared with \texttt{o1}, whose curve extends to $N=140$, most other models stop earlier 
    because their performance trends were already evident, making further evaluation at larger $N$ unnecessary.}
    \label{fig:unsat-small-alpha}
  \end{subfigure}
  \caption{(a) Phase transition for random $3$SAT with $N{=}75$ using a CDCL solver and (b) Correct rate of low-$\alpha$ \UNSAT\ CNFs’ prediction across models as $N$ increases.
  }
  \label{fig:combined-phase-unsat}
\end{figure*}

\subsection{Phase Transition in Random $K$-SAT}
\paragraph{Clause density and criticality.}
A standard random $K$-SAT model fixes $N$ and samples $L$ clauses uniformly at random from all $K$-clauses over $N$ variables.
The key control parameter is the \emph{clause density}
\[
\alpha \;=\; \frac{L}{N}.
\]
For $K\!\ge\!3$ there is a sharp satisfiability threshold at a \emph{critical density} $\alpha_c(K)$ in the $n\to\infty$ limit:
for $\alpha \ll \alpha_c(K)$, instances are \SATLabel\ with high probability; for $\alpha \gg \alpha_c(K)$, instances are \UNSAT\ with high probability.
For $K=3$, the empirical/analytic consensus places $\alpha_c(3)\approx 4.26$~\citep{Mitchell:1992,kirkpatrick1994critical}; 
for $K=2$, the threshold is at $\alpha_c(2)=1$ but the problem remains in P~\citep{Aspvall:1979}.
Around $\alpha_c(K)$, typical algorithmic hardness peaks: solvers require many more decisions/conflicts and runtimes balloon—the canonical \emph{easy–hard–easy} phenomenon~\citep{Mitchell:1992}.
For finite $N$, one observes a pseudo-threshold
$\alpha_c(N)$ (e.g., the location of the $50\%$ satisfiability point or the
peak of median solver cost). As $N$ increases, $\alpha_c(N)$ drifts toward the
asymptotic threshold $\alpha_c \approx 4.26$, and the transition sharpens; the
exact finite-size drift (from above or below) depends on the estimator and the
instance generator.

\paragraph{Hardness peak (finite-size view).}
With finite $N$, the transition is smoothed by finite-size effects, yet the same qualitative picture holds. 
Figure~\ref{fig:phase-cdcl-n200} illustrates this for random $3$-SAT with $N{=}75$ under a modern CDCL solver, \rev{namely MiniSat version 2.2\cite{Een:2003}, which we use to check satisfiability for all formulas in this paper}. 
The probability of satisfiability (blue, right axis) remains close to $1$ for small $\alpha$, then undergoes a sharp decline around the critical threshold $\alpha_c \approx 4.26$, and quickly approaches $0$ as $\alpha$ increases further. 
In parallel, the median number of branching decisions (black, left axis) follows the canonical \emph{easy--hard--easy} pattern, reaching its maximum near $\alpha_c$. 
Although finite-size effects smooth the curves, the empirical behavior aligns with the asymptotic threshold phenomenon: the satisfiability transition sharpens and solver hardness concentrates around $\alpha_c$ as $N$ grows larger.

% \subsection{Experiment 1: Phase Transition Behavior of LLMs}
\section{Initial Misapprehension under Traditional Metrics}
\label{sec:misapprehension}

Having established in the introduction that SAT solving serves as a critical benchmark for assessing the reasoning capability of large language models (LLMs), we began our investigation by following the conventional evaluation approach used in both classical SAT solving and prior machine learning studies. Specifically, we aimed to observe whether LLMs exhibit the well-known \emph{phase transition phenomenon}, a hallmark of SAT problems, and whether their performance could be adequately captured using traditional classification metrics such as accuracy, precision, recall, and F1.

\subsection{Motivation and Research Questions} 
Our first experiment seeks to answer two questions:  
\begin{enumerate*}[label=\textbf{RQ}\arabic*, ref=\textbf{RQ}\arabic*, series=RQlisting]
\item \label{Rquestions:RQ1}
Do LLMs demonstrate the same phase transition behavior observed in classical traditional SAT solving?  

\item \label{Rquestions:RQ2}
Can traditional evaluation metrics reliably measure the reasoning ability of LLMs on SAT formulas?  
\end{enumerate*}
In classical SAT solving, it is well-established that for a fixed number of variables $N$, as the clause-to-variable ratio $\alpha$ increases, the satisfiability of formulas undergoes a sharp transition: almost all formulas are satisfiable when $\alpha$ is small, and almost all formulas become unsatisfiable when $\alpha$ is large. Near the critical threshold (approximately $\alpha \approx 4.26$ for 3SAT), the difficulty of solving formulas increases dramatically, requiring significantly more branches (decision points) and conflicts (backtracking steps). Detecting this phenomenon from the results of LLMs
is therefore a natural first step for assessing whether they possess genuine reasoning capability in SAT solving.

\subsection{Experiment 1 Setup.}
\label{sec:exp1-setup}
To test this, we fixed the number of variables to $N = 75$ and systematically varied $\alpha$ across the range:
\[
\alpha \in \{3.5, 4.0, 4.5, 5.0, 5.5\}.
\] 

\revdel{For each $\alpha$, we generated 320 random 3SAT instances in CNF form. All generated formulas were verified using a classical SAT solver to ensure the correctness of their satisfiability labels.} 

\rev{We chose these $\alpha$ values to straddle the classical 3-SAT threshold
($\alpha_c\approx4.26$), thereby covering \SATLabel-dominated, near-critical, and \UNSAT-dominated regimes in a single
controlled experiment. We fixed $N=75$ to keep CNFs within prompt-length limits while still exhibiting a clear
finite-size phase-transition signature under a CDCL solver (Fig.~\ref{fig:combined-phase-unsat}a). For each $\alpha$, we generated 320 random
3SAT formulas in CNF form and determined their satisfiability labels using 
MiniSat. We generated a relatively large number of instances for each $\alpha$ because when $\alpha$ is close to 4.26 it is quite easy for all generated instances to be all \SATLabel\ or \UNSAT. A large sample instance averts this problem. 
}

\begin{lstlisting}[
    style=promptstyle,
    language=Python,
    caption={LLM prompt for 3-CNF solving},
    label={lst:llm-prompt}
]
prompt = f"""You are a SAT logic solver.  
    Please use a step-by-step method to solve the following 3-CNF formula.  
    Finally, output only the following three items, with no extra explanation:
    * Whether the formula is SATISFIABLE or UNSATISFIABLE  
    * Number of branches (i.e., decision points)
    * Number of conflicts (i.e., backtracking steps)
    If the formula is SATISFIABLE, please give me the value for each variable.
    The formula is:
    {cnf_text} """
\end{lstlisting}

\subsection{LLM Prompting Strategy} 
Each CNF formula was provided to the LLMs through a carefully structured prompt designed to elicit both the satisfiability decision and intermediate reasoning statistics (branches and conflicts). An example prompt is shown in Listing~\ref{lst:llm-prompt}.  
The design intentionally mirrors the behavior of algorithmic solvers, asking the model not only for the final \SATLabel/\UNSAT\ classification but also for the reasoning trace, thereby enabling us to compare LLMs’ output with 
the behavior of algorithmic solvers.

\begin{figure}[t]
  \centering
  \includegraphics[width=\linewidth]{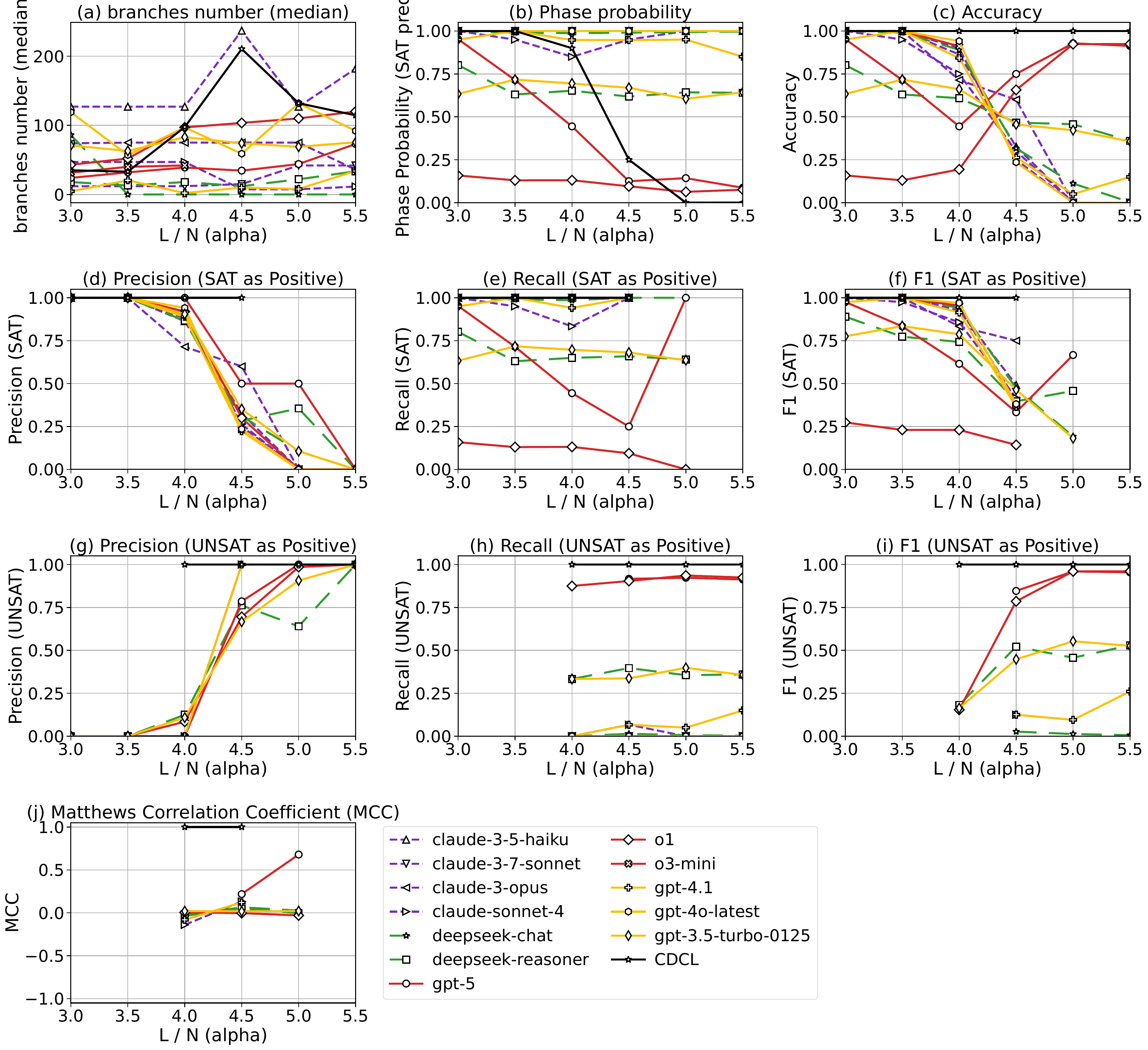}
\caption{
\textbf{Comparison of LLM performance across ten metrics.}
\textbf{1)} Figures in row 2 use \SATLabel\ as positive and those 
in row 3 use \UNSAT\ as positive. This makes sure that all data points are presented as $\alpha$ changes from very small (when almost all instances are \SATLabel) to large (when almost all instances are \UNSAT). \\
\textbf{2)} \emph{Color/line families}: \textbf{(i)} purple dashed = Anthropic \emph{Claude} (4 variants); 
\textbf{(ii)} green dashed = \emph{DeepSeek} (2 variants);
\textbf{(iii)} yellow solid = OpenAI non-reasoning models;
\textbf{(iv)} red solid = OpenAI reasoning-oriented models.
Different marker shapes distinguish individual models within each family.
\textbf{3)}Missing points occur when metrics become undefined, e.g., under highly
    imbalanced data distributions or when a model consistently
    predicts SAT, leading to zero denominators in metric formulas.}
\label{fig:llm-all-metrics}
\end{figure}

\subsection{Observed Outcomes under Traditional Metrics}

\paragraph{Compare median branches} As shown in Figure ~\ref{fig:llm-all-metrics}(a), median branches of CDCL (black curve), 
an algorithmic \SATName\ solver, follow an easy–hard–easy pattern with a
sharp peak near the critical threshold.
In contrast, LLM-reported median branches, as shown in Figure~\ref{fig:llm-all-metrics}(a), do not reproduce the easy–hard–easy pattern. Across LLM models, branch counts remain low or fluctuate idiosyncratically
without the expected spike at $\alpha \approx 4.26$. Notably, \texttt{claude-3-5-haiku} is
the only model that shows a pronounced increase, but its phase probability curve in Figure ~\ref{fig:llm-all-metrics}(b) does not work well. This suggests that the “branches” emitted by LLMs do
not reflect solver-like search depth; they are largely artifacts of generated
reasoning traces rather than an internal backtracking process.
\paragraph{Compare \SATLabel\ probability} As shown in Figure ~\ref{fig:llm-all-metrics}(b), the black curve (CDCL) exhibits the classical phase transition: the fraction of
instances predicted SAT is near $1$ for small $\alpha$, then drops sharply
around the threshold (toward $0$ as $\alpha$ increases). In contrast, most LLMs
produce relatively flat, high curves indicative of a strong \SATLabel\ bias. Most LLM models produce a \emph{flattened or
noisy} curve that stays high even where classical CDCL analysis deems almost all
instances \UNSAT. Notably, the GPT-5 curve (red solid line with small circle markers) is the only model that qualitatively follows the
CDCL trend—a monotone decrease with $\alpha$—but its drop is much more gradual
and occurs later, failing to reproduce the CDCL-like cliff.

\paragraph{Answer to \ref{Rquestions:RQ1}.}
Taken together, these two comparisons indicate that current LLMs do \emph{not}
demonstrate the classical traditional phase-transition behavior. While \texttt{o1} is
relatively better—showing some sensitivity to $\alpha$—it still fails to align
with the expected threshold and does not reproduce the canonical shape.

\paragraph{Compare traditional metrics.}
After examining Figure~\ref{fig:llm-all-metrics}, panels (c) to (i), we observe:

1) Accuracy (Figure~\ref{fig:llm-all-metrics} (c)).
When \(\alpha \le 3.5\), most LLMs reach high accuracy, many at or near \(1.0\), with the exception of \texttt{o1}.
When \(\alpha \ge 5.0\), accuracy for most models drops toward \(0\); the two clear exceptions are \texttt{GPT-5} and \texttt{o1}, which remain substantially higher than the rest.

2) Set \SATLabel\ as the positive class (Figure~\ref{fig:llm-all-metrics} panels (d) to (f)).
For \(\alpha \le 3.5\), almost all LLMs achieve precision \(=1\) and recall and F1 close to \(1\), except \texttt{deepseek-reasoner}, \texttt{gpt-3.5-turbo-0125}, and \texttt{GPT-5}.
For \(\alpha \ge 5.0\), precision of almost all models collapses to \(0\) or near \(0\), with \texttt{GPT-5} and deepseek-reasoner as clear outliers that retain nontrivial precision.
In these bins at high \(\alpha\) many entries of recall and F1 are missing, because the bin contains only \UNSAT\ instances, which makes these two measures undefined.

3) Set \UNSAT\ as the positive class (Figure~\ref{fig:llm-all-metrics} panels (g) to (i)).
For \(\alpha \le 3.5\), precision is \(0\) for all models, and recall and F1 are undefined (thus blank) because these bins contain only SAT instances.
Once \(\alpha \ge 4.0\), \texttt{GPT-5} and \texttt{o1} obtain relatively strong recall and F1;
\texttt{deepseek-reasoner} and \texttt{gpt-3.5-turbo-0125} show F1 values fluctuating around \(0.5\);
the remaining models stay close to \(0\).

4) Based on the performance of the models across different metrics, we conclude that except for \texttt{GPT-5}, \texttt{o1}, \texttt{deepseek-reasoner}, and \texttt{gpt-3.5-turbo-0125}, most LLMs exhibit a strong bias toward predicting SAT. Among all models, \texttt{GPT-5} achieves the best overall performance; the next tier is formed by \texttt{o1}, \texttt{deepseek-reasoner}, and \texttt{gpt-3.5-turbo-0125}. Within this group, \texttt{o1} tends to predict UNSAT more frequently than the others. These trends are consistent with the class imbalance across \(\alpha\) and indicate that high scores in certain bins reflect distributional effects rather than robust logical reasoning.

% \paragraph{Compare MCC.}
% Inspecting Figure~\ref{fig:llm-all-metrics}(j), we observe that the Matthews Correlation Coefficient (MCC)~\citep{matthews1975mcc,chicco2020mcc} 
% is missing at many \(\alpha\) values. This is expected: across clause density \(\alpha\), the \SATLabel/\UNSAT\ distribution is highly imbalanced; at small \(\alpha\) the bins often contain almost only \SATLabel\ instances, whereas at large \(\alpha\) the bins often contain almost only \UNSAT\ instances.
% With a limited
% number of instances per bin, it is common that a bin collapses to a single
% class. In these degenerate cases the MCC is \emph{undefined}, so we record it
% as \texttt{NaN} and the corresponding marker is left blank in the plot.

% \[
% \mathrm{MCC}
% \;=\;
% \frac{TP\cdot TN - FP\cdot FN}
% {\sqrt{(TP+FP)\,(TP+FN)\,(TN+FP)\,(TN+FN)}}.
% \]

% The denominator is a product of four non-negative terms. If any term is zero,
% MCC is undefined. This occurs whenever a bin has only one class in the ground
% truth or the model predicts only one class:
% \begin{itemize}
%   \item All ground truths are \UNSAT\ (no positives): \(TP+FN = 0\).
%   \item All ground truths are \SATLabel\ (no negatives): \(TN+FP = 0\).
%   \item The model never predicts \SATLabel\ (always \UNSAT): \(TP+FP = 0\).
%   \item The model never predicts \UNSAT\ (always \SATLabel): \(TN+FN = 0\).
% \end{itemize}

% \noindent
% Therefore, MCC is only defined (and thus plotted) at \(\alpha\) where both the
% ground truth and the predictions contain \emph{both} classes within the bin.

\paragraph{Compare MCC}
Figure~\ref{fig:llm-all-metrics}(j) shows that the Matthews
Correlation Coefficient (MCC)~\citep{matthews1975mcc,chicco2020mcc},
\[
\mathrm{MCC}
=\frac{TP\cdot TN - FP\cdot FN}
{\sqrt{(TP+FP)(TP+FN)(TN+FP)(TN+FN)}},
\]
is missing at many \(\alpha\) values, because the imbalanced \SATLabel/\UNSAT\
distribution causes per-bin ground truth or predictions to collapse to a single
class, zeroing one denominator term and leaving MCC \emph{undefined}. We provide
the full enumeration in the supplemental material, section
``Degenerate Cases of MCC under Class Imbalance''.

\paragraph{Answer to \ref{Rquestions:RQ2}.}
These observations expose the \emph{initial misapprehension} behind applying
traditional classification metrics to LLM-based SAT solving. Accuracy,
precision, recall, and F1 are confounded by the $\alpha$-dependent label
imbalance and cannot reliably measure reasoning ability. Even when a model such
as \texttt{o1} numerically outperforms its peers under these metrics, the score
improvements are not indicators of solver-like reasoning because the phase
transition signature is missing.

\smallskip
\noindent
In summary, LLM outputs do not replicate the classical 
phase transition (\ref{Rquestions:RQ1}: \emph{no}), and conventional metrics fail to provide a
trustworthy assessment of reasoning ability in this setting (\ref{Rquestions:RQ2}: \emph{no}).
This motivates our pair-balanced evaluation via \emph{paired formulas}
in Section~\ref{sec:paired-formulas}.

\subsection{Limitations and Lessons Learned}  

Our findings reveal several fundamental limitations when evaluating LLMs’ SAT-solving capability under traditional metrics. These limitations stem from both the inherent properties of SAT distributions and the biases of current LLMs.
% , which together undermine the validity of conventional evaluation methods such as accuracy, precision, recall, and F1. 
We summarize two major issues below.  

\paragraph{(1) Imbalanced \SATLabel/\UNSAT\ distributions across $\alpha$.}  
As established in classical traditional SAT theory, the satisfiability distribution shifts dramatically with the $\alpha$ clause density~\citep{Mitchell:1992,kirkpatrick1994critical}: when $\alpha$ is low, nearly all generated formulas are satisfiable; when $\alpha$ is high, nearly all become unsatisfiable. This imbalance means that trivial baselines (e.g., always predicting \SATLabel\ at low $\alpha$ or \UNSAT\ at high $\alpha$) can achieve deceptively high accuracy and recall. Consequently, traditional metrics cannot disentangle genuine reasoning capability from simple exploitation of label imbalance.  

\paragraph{(2) LLMs’ strong bias toward predicting \SATLabel}  
Across models, we observed a systematic bias in favor of \SATLabel\ predictions, regardless of $\alpha$. This bias is evident in the “flattened” SAT-probability curves (Figure~\ref{fig:llm-all-metrics} (b)), which stay high even in regimes where virtually all instances are unsatisfiable. The bias exacerbates the label imbalance issue and further distorts metrics such as precision, recall, and F1. Although \texttt{o1} departs somewhat from this trivial behavior, its performance still falls far short of solver-like fidelity.

\section{Sensitivity to Problem Size and UNSAT Detection}
\label{sec:unsat-detection}

\subsection{Research Question}
As discussed in Section~\ref{sec:misapprehension}, our initial evaluation revealed two limitations of applying traditional metrics to LLM-based SAT solving.
% : (1) imbalanced distributions of SAT versus UNSAT instances across $\alpha$, (2) a strong bias of LLMs toward predicting SAT, and (3) a lack of sensitivity to the effect of problem size $N$.  
In addition, $N$ was fixed to 75, and thus it was not clear
how problem size influences LLM behavior. 
Without varying $N$, conclusions about reasoning ability remain incomplete.
% and traditional metrics may mask weaknesses that only emerge at larger instance sizes.  
%
These motivate the following research question:
\begin{enumerate*}[resume*=RQlisting]
\item \label{Rquestions:RQ3}

Can LLMs reliably detect \UNSAT\ formulas where \SATLabel\ dominate, and how does scaling the problem size $N$ influence this ability?
\end{enumerate*}

\subsection{Experiment 2 Setup}
\label{sec:exp2-setup}

To directly address \ref{Rquestions:RQ3}, we deliberately constructed more \UNSAT\ formulas at low $\alpha$, thus isolating LLMs’ ability to detect unsatisfiability. At the same time, we varied the number of variables $N$ to examine the effect of scaling problem size.
Specifically, we varied $N$ across thirteen settings. 
\[
N \in \{5, 8, 10, 25, 50, 60, 70, 80, 90, 100, 110, 120, 140\}
\]

\revdel{For each $N$, we generated 70 random 3SAT formulas, each verified as \UNSAT\ by CDCL.}
\rev{We varied $N$ over a wide grid up to 140 to expose scale effects while keeping the per-$N$ sample size feasible
(70 instances) under API and computation constraints. To explicitly stress \UNSAT\ detection in a low-$\alpha$ regime, each generated instance
has an $\alpha \in \{3.5, 3.6, 3.7, 3.8, 3.9, 4.0\}$. Also, we discarded all instances whose labels are \SATLabel\ based on a CDCL solver.  
For each $N$ we generated 70 instances in this manner, 
We then queried each model with the same prompt as in Experiment~1.} 

\rev{For some
models we stop at smaller $N$ once trends are monotone and stable (Fig.~\ref{fig:combined-phase-unsat}b) to
reserve budget to paired evaluation and cross-representation tests later. While this reaches $N=140$ for selected models, fully charting the failure boundary at even larger $N$ (or under harder instance distributions) remains constrained by prompt-length limits, timeouts, and API budget; we therefore treat this experiment as a scale stress-test and discuss practical scaling strategies in Sec.~\ref{sec:discussion}.}

\revdel{The clause-to-variable ratio $\alpha$ for each formula was chosen randomly from:
\[
\alpha \in \{3.5, 3.6, 3.7, 3.8, 3.9, 4.0\}.
\]
The same structured prompt as in Experiment~1 in Section~\ref {sec:misapprehension} was used to query each LLM.}

\subsection{Results}
Figure~\ref{fig:unsat-small-alpha} reports the correct prediction rates of low-$\alpha$ \UNSAT\ CNFs as $N$ increases. Several trends answer \ref{Rquestions:RQ3}:

1) Negative correlation with problem size. The correct prediction rate declines with larger $N$, with sharp drops once $N \geq 25$. This shows that scaling remains a fundamental barrier.  

2) Reasoning-oriented models perform better. Models such as  \texttt{GPT-5}, \texttt{o1}, \texttt{o3-mini}, and deepseek-reasoner consistently outperform general-purpose models. Although their performance also deteriorates, they maintain relatively higher \UNSAT\ detection rates.  

3) Persistent \SATLabel\ bias in non-reasoning models. Many general-purpose LLMs (e.g., \texttt{gpt-3.5-turbo}, \texttt{claude-3.5-haiku}) fail almost entirely once $N$ increases, confirming their strong bias toward \SATLabel.

% \begin{figure}[t]
%   \centering
%   \includegraphics[width=.9\linewidth]{figures/unsat/unsat_small_alpha_prediction_correct_rate.pdf}
%   \caption{Correct rate of low-$\alpha$ \UNSAT\ CNFs’ prediction across models as $N$ increases. Compared with \texttt{o1}, whose curve extends to $N=140$, most other models stop earlier 
% because their performance trends were already evident, making further evaluation at larger 
% $N$ unnecessary.}
%   \label{fig:unsat-small-alpha}
% \end{figure}

\subsection{Lessons Learned}
From these results, we derive three findings regarding \ref{Rquestions:RQ3}
:

1) Most LLMs cannot reliably detect \UNSAT\ formulas when \SATLabel\ dominates the distribution.  

2) Reasoning-oriented models provide measurable improvements but still degrade significantly with larger $N$.  

3) Stress-testing with targeted \UNSAT\ examples is essential, as traditional metrics over random distributions cannot reveal these weaknesses.  

Thus, the answer to \ref{Rquestions:RQ3} is largely negative: LLMs struggle to detect \UNSAT\ formulas at scale, and although reasoning models perform better, they remain limited. This validates the need for distribution-aware and problem-size-sensitive evaluations, and sets the stage for moving beyond aggregate metrics toward paired-formula testing.

\section{Refined Understanding with Paired Formulas}
\label{sec:paired-formulas}

\subsection{Motivation}
As observed in Experiment~1 in Section~\ref{sec:misapprehension}, randomly generated CNFs at the same $\alpha$ exhibit highly imbalanced \SATLabel\ versus \UNSAT\ distributions. This imbalance undermines traditional metrics such as accuracy, precision, and recall, since models can exploit label skew rather than possess genuine reasoning to achieve high scores.

These limitations point to a deeper need: an evaluation framework that is both distribution independent and sensitive to minimal structural changes in formulas. To this end, we introduce \emph{paired formulas}—minimally different \SATLabel/\UNSAT\ counterparts—and a new evaluation measure, the \emph{Accurate Differentiation Rate (ADR)}, which directly tests whether models can classify both members of a pair correctly.  

% From a theoretical perspective, 
Essentially,
paired formulas eliminate superficial statistical cues and force the model to rely on genuine logical reasoning. Each pair differs by a minimal but logically critical
edit (e.g., flipping a literal), so the only way to succeed consistently is to capture the structural shift that collapses or preserves the solution space. ADR operationalizes this by requiring correct classification of both members of each pair, thus serving as a direct proxy for reasoning sensitivity. In this sense, ADR is not merely another accuracy-like metric: it quantifies whether the model truly tracks constraint propagation and logical consistency, rather than exploiting spurious correlations.  

Together, paired formulas and ADR raise the following question:  
\begin{enumerate*}[resume*=RQlisting]
\item \label{Rquestions:RQ4}

Can paired formulas and ADR provide a more reliable evaluation of LLMs’ reasoning ability in SAT solving?
\end{enumerate*}

\subsection{Experiment 3 Setup}
\revdel{For each \UNSAT\ formula generated 
in Section~\ref{sec:unsat-detection}, we constructed a corresponding 
\SATLabel\ formula via a \emph{minimal adjustment}, designed to preserve the same number of variables $N$ while keeping the clause-to-variable ratio $\alpha$ nearly unchanged. Adjustments were deliberately crafted to introduce only the smallest possible modifications, such as:}
\rev{For each solver-verified \UNSAT\ formula $F^{\textsf{UNSAT}}$ from Sec.~\ref{sec:unsat-detection}, we generate a paired
\SATLabel\ counterpart $F^{\textsf{SAT}}$ via a \emph{single-edit} procedure, in the following order.} 
\begin{enumerate}
\item \revdel{deleting a single clause,}  \rev{flip the polarity of one literal in a randomly selected clause such as changing $x$ to $\neg x$ or vice versa;}
\item \revdel{flipping the polarity of a literal, or}  \rev{replace one literal with another variable while preserving clause width;}
\item \revdel{replacing one literal with another in the same position.}  \rev{delete one clause.}
\end{enumerate}
\rev{The edit procedure stopped once an \SATLabel\ instance had been created (verified by MiniSat), which we succeeded for each \UNSAT\ instance within these three edits. This procedure kept the variable set fixed and the $\alpha$ values for the paired instances differ to be at most by $\frac{1}{N}$. The full editing script and edit statistics are given in the supplemental material, which is included in the supplementary package linked from the ``Data Availability'' Section.} 

Each \UNSAT\ formula $F^{\text{UNSAT}}$ was thus paired with a structurally near-identical \SATLabel\ formula $F^{\text{SAT}}$. The generated dataset consists of balanced pairs differing only by minimal but \emph{logically distinct} changes, ensuring that evaluation tested genuine reasoning ability rather than the reliance on superficial heuristics. 

\subsection{Evaluation Methodology}
The same structured prompt was used as in previous experiments. Since the dataset was balanced across pairs, we computed traditional metrics (accuracy, precision, recall, F1) under fair conditions.  

In addition, we proposed the \emph{Accurate Differentiation Rate} (ADR), defined as:
\[
\mathrm{ADR} = \frac{1}{M} \sum_{i=1}^M \mathbf{1}\Big(
    \hat{y}(F^{\text{SAT}}_i) = \text{\SATLabel} \ \wedge\
    \hat{y}(F^{\text{UNSAT}}_i) = \text{\UNSAT}
\Big).
\]
ADR measures whether a model correctly classifies \emph{both} members of a pair. Unlike accuracy or recall, which remain sensitive to prediction bias, ADR directly evaluates whether models can differentiate formulas that are structurally similar but logically distinct. Intuitively, high ADR indicates that a model performs constraint-level reasoning akin to a SAT solver, whereas low ADR suggests reliance on shallow heuristics.

\begin{figure}[t]
  \centering
  \includegraphics[width=\linewidth]{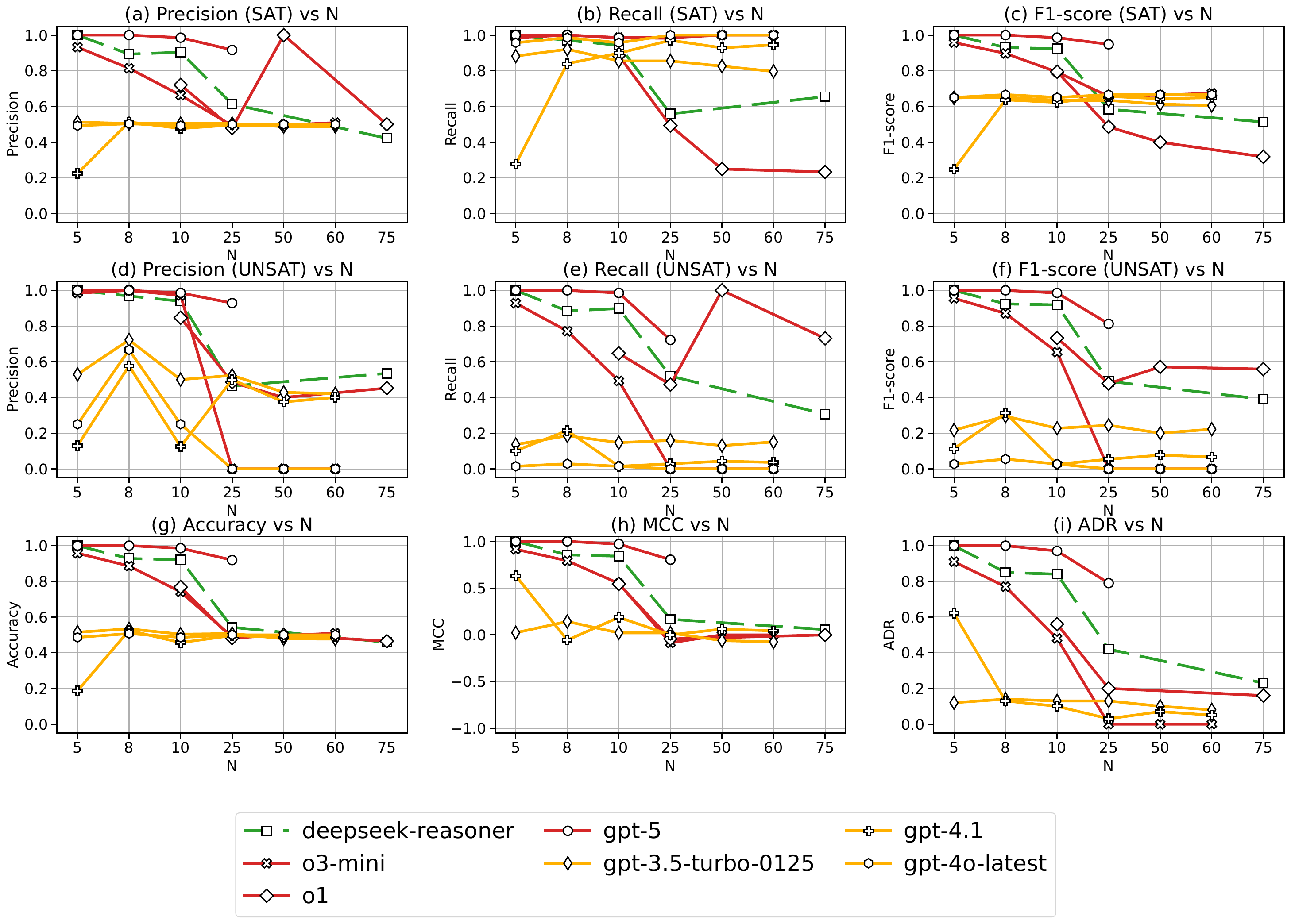}
  \caption{
  Performance of different models on paired \SATLabel/\UNSAT\ CNF instances across multiple metrics. 
\textbf{1)} Panels (a)--(h) report traditional classification scores (precision, recall, F1, accuracy, MCC) for \SATLabel\ and \UNSAT\ separately. 
  % While these metrics appear near-perfect at small $N$ and degrade inconsistently as $N$ grows, they are strongly confounded by class imbalance and trivial guessing strategies (e.g., always predicting \SATLabel\ yields high recall). 
\textbf{2)} Panel (i) shows the proposed \textbf{Accurate Differentiation Rate (ADR)}, which directly measures the fraction of pairs where both \SATLabel\ and \UNSAT\ predictions are correct. 
  Unlike conventional metrics, ADR declines smoothly with increasing $N$, revealing the true collapse of reasoning ability and providing a clearer, assessment of model competence.
\textbf{3)} \emph{Experiment scope:} \texttt{GPT-5} results stop at $N=25$ due to frequent timeouts at larger $N$ \rev{(which yielded low completion rate), not because ADR ceases to apply. ADR remains well-defined beyond $N=25$ (e.g., Fig.~3(i) shows meaningful trajectories for \texttt{deepseek-reasoner} and \texttt{o1} at larger $N$). In an additional validation run, \texttt{GPT-5} achieved ADR $=0.35$ at $N=50$ (down from $0.79$ at $N=25$), but this point was omitted from the plot because the $N=50$ completion rate was only about $20\%$ (most runs timed out).} 
\texttt{o1} starts from $N \geq 10$ given its validated performance at smaller $N$ and higher API costs; 
\texttt{o1} and \texttt{deepseek-reasoner} were run at $N=75$ (but not $N=60$), 
whereas the other models were run at $N=60$ (but not $N=75$), primarily due to budget constraints.}
\label{fig:pairs-small-alpha-metrics}
\label{fig:paired-varying-N}
\end{figure}

\subsection{Comparative Analysis: Discussion of Results}

In this section we contrast the proposed \emph{Accurate Differentiation Rate (ADR)} with conventional classification metrics, including accuracy, precision, recall, F1-score, and Matthews correlation coefficient (MCC). Our comparison is based on the paired CNF experiments with varying $N$s and for each $N$ we generated 70 \SATLabel/\UNSAT\ pairs.

\subsubsection{Limitations of Traditional Metrics}
Figures~\ref{fig:paired-varying-N}(a)--(h) show that traditional metrics are highly sensitive to class-specific biases and label distributions. For small $N$ (e.g., 5 or 10), all models achieve near-perfect precision, recall, and F1 on \SATLabel\ instances, while their performance on \UNSAT\ instances varies widely. This asymmetry artificially inflates overall accuracy, giving the impression of robust reasoning where models are in fact exploiting superficial biases (e.g., over-predicting \SATLabel). 

Moreover, MCC values (Figure~\ref{fig:paired-varying-N}(h)) become negative for larger $N$, reflecting inconsistent predictions across classes. While mathematically sound, such negative correlations are difficult to interpret in terms of reasoning ability. In sum, conventional metrics are easily confounded: high recall on \SATLabel\ can coexist with trivial guessing strategies, while accuracy (Figure~\ref{fig:paired-varying-N} (g)) and MCC (Figure~\ref{fig:paired-varying-N} (h)) provide contradictory signals under class imbalance.

\subsubsection{Advantages of ADR}
By design, ADR singles out the joint-correctness signal: it measures the fraction of \SATLabel/\UNSAT\ pairs where the model predicts both formulas correctly. As shown in Figure~\ref{fig:paired-varying-N}(i), ADR aligns with the true reasoning difficulty of the task. For small $N$, ADR is close to~1, consistent with the models' ability to handle simple cases. As $N$ grows, ADR decreases smoothly and monotonically, revealing the exact point at which reasoning collapses. 

Compared with \textit{Accuracy}, when \(N \ge 25\) the accuracies of all models
(except \texttt{GPT-5}) cluster around \(0.5\).
This does \emph{not} indicate a reasoning ability of \(0.5\), because with two
classes a random guesser already attains \(0.5\) accuracy.
By contrast, \textit{ADR} faithfully captures the collapse of reasoning for
\(N \ge 25\), with scores often dropping below \(0.2\).
Moreover, the cross-model ordering—reasoning-oriented models (red/blue curves)
lying above non-reasoning ones (yellow)—further shows that reasoning models
outperform non-reasoning models. 
Unlike MCC, ADR never yields negative or ambiguous values, and unlike per-class precision/recall, ADR is invariant to label asymmetries.

\subsubsection{Summary}
ADR provides a cleaner view of reasoning ability in paired evaluations. Traditional metrics can be inflated by trivial guessing or class imbalance, whereas ADR directly quantifies whether a model can consistently differentiate satisfiable from unsatisfiable counterparts. The experimental evidence confirms that ADR is a more faithful indicator of reasoning competence, especially as problem size increases.

\subsection{Lessons Learned}
% Experiment~3 shows 
Previous subsections show 
that paired formulas and ADR provide a more faithful probe of LLM reasoning ability. By controlling for label imbalance and testing sensitivity to minimal structural changes, this methodology reveals whether predictions reflect true logical understanding rather than heuristics.  

From a theoretical standpoint, paired formulas function as a “reasoning stress test”: they remove superficial cues and expose whether a model can track the logical consequences of minimal constraint edits. ADR then quantifies this sensitivity, serving as an observable proxy for reasoning strength.

Thus, the answer to \ref{Rquestions:RQ4} is affirmative: while traditional metrics remain confounded by bias, ADR offers a direct window into logical reasoning. Paired formulas and ADR refine our understanding of LLM performance and establish a stronger foundation for future evaluations. Using paired formulas, ADR isolates joint correctness and tracks reasoning difficulty: it is near~1 at small $N$ but drops smoothly with size—non-reasoning models typically fall below~0.2 by $N\!\ge\!25$ while accuracies hover around~0.5 (i.e., random guessing) (Figure~\ref{fig:paired-varying-N}(i)). 
Thus, current LLMs do not reliably solve SAT at scale; traditional metrics are confounded by label bias, whereas ADR offers a faithful assessment.

\subsection{Mathematical Properties of ADR}
\label{sec:adr-math}

We formalize the \emph{Accurate Differentiation Rate (ADR)} in the paired-evaluation setting. 
Let 
\[
\{(F_i^{\text{SAT}},\, F_i^{\text{UNSAT}})\}_{i=1}^M
\]
be $M$ minimally-different pairs and define
\[
A_i = \mathbf{1}\{\hat{y}(F_i^{\text{SAT}})=\text{\SATLabel}\},\qquad
B_i = \mathbf{1}\{\hat{y}(F_i^{\text{UNSAT}})=\text{\UNSAT}\}.
\]

Per-class recalls and accuracy are
\[
r_S=\frac{1}{M}\sum_{i=1}^M A_i,\quad
r_U=\frac{1}{M}\sum_{i=1}^M B_i,\quad
\mathrm{Acc}=\tfrac{1}{2}(r_S+r_U).
\]
ADR is the \emph{joint-correct} rate over pairs:
\[
\mathrm{ADR}=\frac{1}{M}\sum_{i=1}^M A_iB_i.
\]

\paragraph{Tight bounds and consequences.}
ADR satisfies the inequality chain
\[
\max(0,\,r_S{+}r_U{-}1)\;\le\;\mathrm{ADR}\;\le\;\min(r_S,r_U)\;\le\;\mathrm{Acc}{=}\tfrac{1}{2}(r_S{+}r_U).
\]
Thus ADR is \emph{stricter} than accuracy/F1: it requires \emph{simultaneous} correctness on both members of a pair and is therefore robust across clause density $\alpha$~\citep{powers2011evaluation,manning2008ir}.

\paragraph{Independence heuristic.}
Let $A=\{F^{\text{SAT}}\text{ correct}\}$ and $B=\{F^{\text{UNSAT}}\text{ correct}\}$, then 
\[
\mathrm{ADR}=P(A\cap B)=r_Sr_U+\mathrm{Cov}(\mathbf{1}_A,\mathbf{1}_B).
\]
, where $\mathrm{Cov}(\mathbf{1}_A,\mathbf{1}_B)$ denotes the covariance of the two indicator variables across pairs.
If $A$ and $B$ were independent, $\mathrm{ADR}=r_Sr_U$; deviations indicate positive/negative coupling between the two decisions.

\paragraph{Intuitive examples.}
(1) \emph{Always predict \SATLabel:} $(r_S,r_U)=(1,0)$ gives $\mathrm{Acc}=0.5$ but $\mathrm{ADR}=0$.  
(2) \emph{One side perfect, one side moderate:} $r_S=1.0,\,r_U=0.6$ implies $\mathrm{ADR}=0.6$ while $\mathrm{Acc}=0.8$.  
In both examples, ADR exposes the true \emph{pairwise} limitation that is masked by accuracy/F1.

\medskip
\noindent\textit{Summary.} ADR cleanly quantifies whether minimal, satisfiability-flipping edits are \emph{jointly} tracked. A detailed derivation of ADR from a mathematical perspective is provided in \revdel{Appendix}\rev{supplemental material}, entitled Mathematical Properties of ADR.

\subsection{ADR vs.\ MCC: A Pair-Level View}
\label{sec:adr-vs-mcc}
Let $n_{11},n_{10},n_{01},n_{00}$ count pairs that are both-correct,
only-\SATLabel-correct, only-\UNSAT-correct, and both-wrong, and let
$M$ be the total. Then $\mathrm{ADR}=n_{11}/M$, while MCC mixes $n_{11}$
with the both-wrong mass $n_{00}/M$ and the class asymmetry
$(n_{10}-n_{01})/M$, and is undefined under class
collapse~\citep{matthews1975mcc,chicco2020mcc}. ADR isolates the
pairwise joint-correct component and is always defined; two models can
share the same MCC yet differ markedly in ADR. We thus use ADR as the
primary diagnostic and MCC as a secondary, distribution-sensitive
summary; full derivations appear in the supplement, section
``ADR vs.\ MCC: A Mathematical Comparison''.

\section{2-SAT as a Controlled Baseline}
\label{sec:twosat}

\subsection{Motivation}
As argued in Section~\ref{sec:Intro}, SAT is our primary probe of LLM reasoning. While 3SAT is NP-complete and stresses models at modest sizes, \emph{2-SAT} lies in~P and removes much of the algorithmic hardness. Evaluating 2-SAT next to 3SAT therefore lets us disentangle \emph{reasoning competence} from \emph{problem difficulty} and observe how performance scales with~$N$ under a cleaner logical structure.  

This motivates the following question:
\begin{enumerate*}[resume*=RQlisting]
\item \label{Rquestions:RQ5}

How does evaluating LLMs on 2-SAT, a polynomial-time fragment of SAT, refine our understanding of their reasoning competence compared to 3-SAT, particularly under paired evaluation with ADR?
\end{enumerate*}

\subsection{Experiment 4 Setup}
We construct \textbf{70 paired} 2-SAT instances (\SATLabel/\UNSAT) per setting, with $N \in \{5, 8, 10, 25, 50\}$. Each pair differs by a single satisfiability-changing edit (e.g., literal polarity flip or one-clause tweak), ensuring balanced labels and minimizing superficial cues. We reuse the structured prompts from the 3SAT experiments; models output a categorical decision and, if predicting \SATLabel, an assignment to variables.

\begin{figure}[t]
  \centering
  \includegraphics[width=\linewidth]{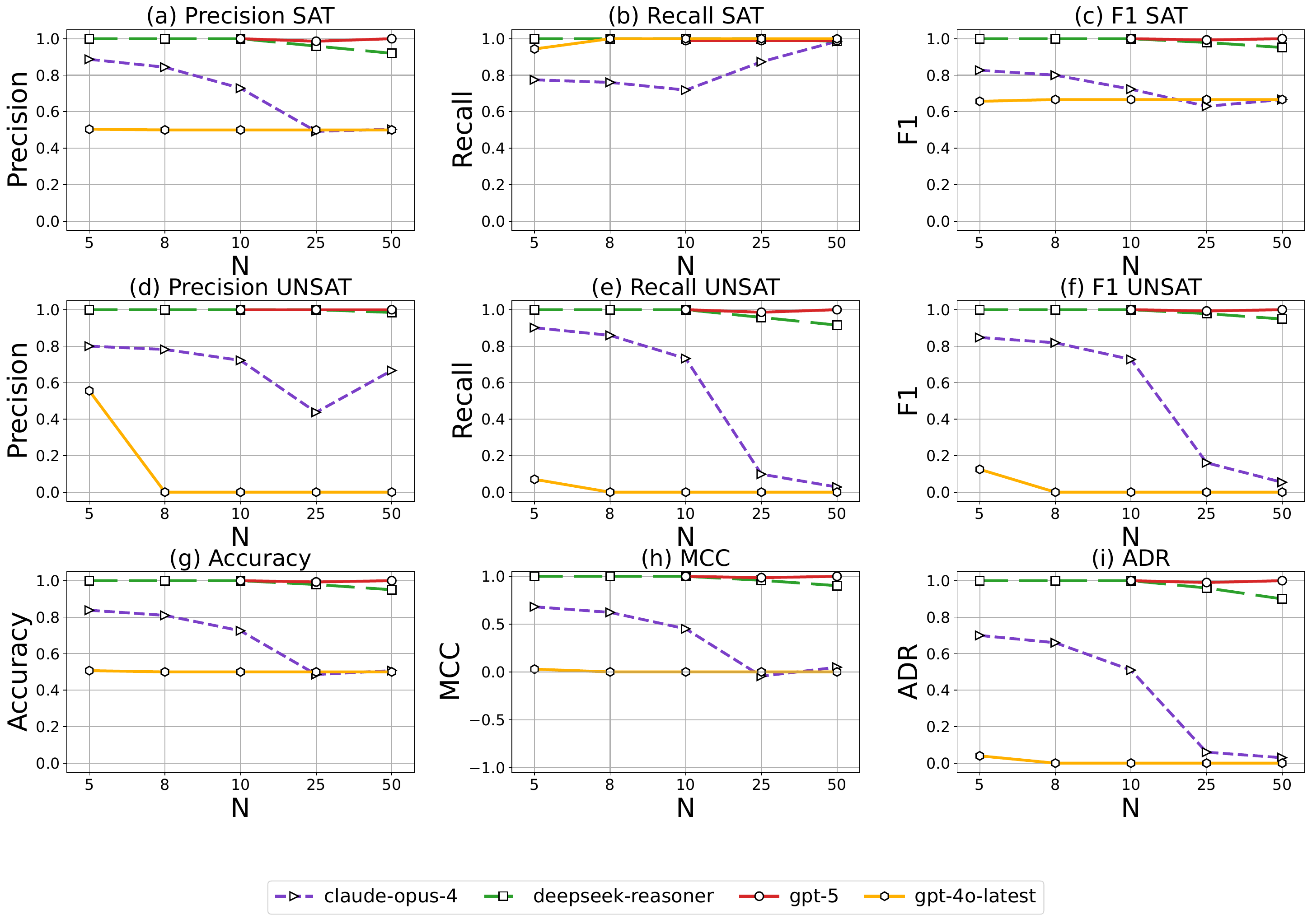}
  \caption{Performance comparison on 2-SAT instances using Precision, Recall, F1, Accuracy, MCC and ADR metrics. GPT-5 was not evaluated for \(N<10\).
Earlier experiments already showed strong performance in this small \(N\) regime,
so we omitted these settings to conserve API budget.}
  \label{fig:2sat-results}
\end{figure}

\subsection{Analysis of Results}
Figure~\ref{fig:2sat-results} reveals several important patterns. First, the reduced complexity of 2-SAT lifts conventional metrics at small $N$: both GPT-5 and deepseek-reasoner achieve near-perfect precision, recall, and F1 on \SATLabel\ cases (Figure~\ref{fig:2sat-results} (a--c)), as well as \UNSAT\ precision (Figure~\ref{fig:2sat-results}) (d) and overall accuracy (Figure~\ref{fig:2sat-results} (g)). In contrast, claude-opus-4 already shows degradation, confirming that 2-SAT is a simpler fragment but
does not erase reasoning requirements. As $N$ increases, gaps widen. GPT-5 sustains strong performance across all metrics, deepseek-reasoner becomes unstable at $N{=}25$, and claude-opus-4 collapses sharply, with recall and F1 for \UNSAT\ instances (Figure~\ref{fig:2sat-results} (e--f)) dropping toward zero, reflecting a persistent SAT bias also observed in 3-SAT.  

Second, MCC (Figure~\ref{fig:2sat-results} (h)) corroborates these findings: GPT-5 and deepseek-reasoner retain high correlation with ground-truth labels, while claude-opus-4 hovers near zero, and even negative values emerge as $N$ grows. This alignment across ADR and MCC underscores that pairwise and correlation-based metrics are more diagnostic of genuine reasoning than raw classification scores.

Finally, ADR (Figure~\ref{fig:2sat-results} (i)) provides a sharper lens than Accuracy or F1. While traditional metrics can appear inflated by one-sided correctness, ADR requires simultaneous correctness on both members of a pair. As a result, ADR clearly separates GPT-5, which remains highly consistent, from deepseek-reasoner, which dips at medium scales, and from claude-opus-4, whose pairwise reasoning fails almost entirely beyond $N{=}10$.

In summary, 2-SAT validates that lower problem complexity temporarily boosts all models, but scaling still exposes structural weaknesses. ADR and MCC remain the most faithful probes of logical competence, aligning with assignment validity and revealing reasoning robustness that conventional metrics obscure.

\subsection{Lessons Learned}

% Lift at small $N$, divergence under scale.
\parag{Sensitivity to $N$}
Across both settings, easier structure yields strong scores at small sizes.  For 2SAT, most models reach near-perfect precision/recall/F1 and high accuracy when $N\!\le\!10$. For 3SAT the same metrics often look healthy at the smallest $N$ but deteriorate much sooner as $N$ grows. The gap widens with scale: 3SAT exposes sharp breakdowns in \UNSAT\ detection and overall stability, whereas 2SAT delays (but does not remove) this collapse.

\parag{ADR as the faithful indicator}
Traditional metrics can appear inflated by one-sided correctness (e.g., overpredicting \SATLabel). In contrast, \emph{Accurate Differentiation Rate (ADR)} declines smoothly with $N$ in both tasks and cleanly separates models that truly track satisfiability-flipping edits from those that rely on superficial heuristics. For 2-SAT, ADR stays high longer (reflecting lower intrinsic difficulty), but it eventually drops. For 3SAT, the ADR decline begins earlier and is steeper, identifying the true onset of reasoning failure that accuracy/F1 may mask.

\parag{Bias and MCC behavior}
A persistent \SATLabel\ bias is magnified in 3SAT and only partially hidden in 2-SAT. Consequently, MCC may become undefined or volatile in 3-SAT bins with class collapse, and can still be misleading under asymmetric predictions. ADR avoids these pathologies by directly requiring joint correctness on paired instances.

\parag{Model ranking and summary}
Reasoning-oriented models (e.g., \texttt{GPT-5}) maintain top ADR, with 2-SAT showing a higher ceiling and slower decay; others (e.g., \texttt{deepseek-reasoner}) benefit similarly at small $N$ but show instability at intermediate sizes; models with strong \SATLabel\ bias (e.g., \texttt{claude-opus-4}) collapse under scale, particularly on 3-SAT. Overall, 2SAT serves as a \emph{controlled baseline} that lifts raw scores yet preserves the central message: \textbf{paired evaluation with ADR is the most reliable indicator of genuine logical competence, and scaling remains the dominant bottleneck revealed most clearly by 3SAT.}

\section{Vertex Cover: Equivalent or Not?}
\label{sec:vc}

\subsection{Motivation and Setup}
SAT solving is a foundational probe of reasoning, and many NP problems reduce to it. To test whether LLMs reason in a representation-invariant way, we reduced the paired CNF formulas (Section~\ref{sec:paired-formulas}) to \emph{Vertex Cover} instances and asked models to decide feasibility ($|C|\le k$) and, if \texttt{YES}, to output a cover. Full prompt design, datasets, and protocol are in the supplemental material.

\subsection{Key Findings}
Most models perform near chance on vertex cover problems, except for \texttt{GPT-5} (and partly deepseek-reasoner), which achieve strong results on small $N$ but degrade as $N$ increases.  
\rev{We used several ways to compare LLMs' prediction performance for SAT and VC. At the instance level, 75.3\% of all SAT instances and their VC counterparts have the exact same prediction results when all LLMs are accumulated. For those that are different, models are more likely to be correct for SAT (73\%) than for VC (27\%). When measured using ADR, the performance of reasoning models for SAT is much higher than for VC. For example, when $N=25$, GPT-5's ADR for SAT is 0.79 but that for VC is only around 0.27.
Similar performance drop is also observed for \texttt{deepseek-reasoner}.
}

\rev{Overall, this experiment shows that LLM's performance on SAT is much better than on VC.}
Also, ADR again provides the clearest signal: it smoothly declines with $N$ and separates reasoning-oriented models from others, while MCC remains noisy or ambiguous.

% \rev{\subsection{Do reductions yield systematic gains over SAT?}
% \label{sec:upperbound-test}
% High decision agreement across representations does not imply that downstream tasks are easier than SAT; it may
% instead reflect stable heuristics that ignore the witness constraints. To directly test whether reductions provide
% \emph{systematic gains} over the SAT baseline, we compare per-instance correctness on each paired CNF $(F^{\textsf{SAT}},F^{\textsf{UNSAT}})$ against correctness on its reduced instance $R(F)$.
% For each model and each $N$, we compute $\Delta_{\textsf{task}}=\textsf{Acc}_{\textsf{task}}-\textsf{Acc}_{\textsf{SAT}}$ and likewise for ADR.
% Across Vertex Cover and 3D packing, $\Delta_{\textsf{task}}$ is non-positive for the large majority of instances and
% never shows a consistent positive shift across $N$; models that fail on SAT typically also fail after reduction,
% often producing invalid witnesses. We therefore interpret SAT as a conservative probe
% in the \emph{empirical} sense: within our evaluated canonical reductions, models do not outperform their SAT
% baseline.}

% Add a figure placeholder (to be filled with your new plot):

% \rev{
% \begin{figure}[t]
%   \centering
%   % \includegraphics[width=\linewidth]{figs/delta_task_vs_sat.pdf}
%   \caption{(NEW) Per-model distribution of $\Delta_{\textsf{task}}$ (Vertex Cover / 3D packing minus SAT baseline) by
%   $N$. Positive values would indicate systematic gains from the reduction; we observe none.}
%   \label{fig:delta-task}
% \end{figure}
% }

\subsection{Implications}
These findings highlight that:  
(i) SAT invariants transfer to Vertex Cover, supporting its role as a foundational problem;  
(ii) representation invariance is useful only when paired with genuine competence;  
(iii) scalability remains the main barrier; and  
(iv) ADR is the most reliable indicator of reasoning ability across representations.  
Extended details and analysis appear in the \revdel{Appendix}\rev{supplemental material}.

\section{3SAT Transfer to 3D Packing}
\label{sec:packing}

\subsection{Motivation and Setup}
To examine \emph{problem-family transfer} beyond CNF and graphs, we reduced the paired 3SAT instances into a discrete \emph{3D packing} formulation whose feasibility exactly mirrors satisfiability.  
Models were prompted to output both a categorical decision and, when answering \texttt{YES}, a concrete assignment.  
Full reduction rules, prompt design, and examples are provided in the \revdel{Appendix}\rev{supplemental material}.

\subsection{Key Findings}
Our experiment results show that 
most models achieve strong precision/recall/F1 only at small $N$ but quickly degrade with scale; non-reasoning models remain biased toward SAT.  
Crucially, ADR again proves most diagnostic: reasoning models show high ADR on small instances, followed by a smooth decline toward chance as $N$ grows, while accuracy and MCC often look acceptable despite systematic failures.  
Moreover, higher ADR correlates with valid packing witnesses, indicating that pairwise discrimination aligns with producing logically correct outputs.

\rev{We again compared LLMs' performance on SAT and on 3D packing at both the instance and the ADR levels. At the instance level, the ratio that LLMs predicted the same results for 3SAT and their corresponding instances in 3D packing was 80.2\%. When the prediction results were different, LLMs were equally correct for SAT (53\%) and 3D packing (47\%). At the ADR level, the performance of reasoners for SAT is again better than that for 3D packing.}

\subsection{Implications}
The packing task preserves the essential difficulty landscape of SAT and confirms that \textbf{SAT is a foundational probe of LLM reasoning}: models carry both their strengths and weaknesses across equivalent NP formulations.  
Traditional metrics can mislead under class imbalance, but ADR reliably tracks reasoning ability and its scale-induced degradation.  
Detailed construction, metrics, and extended analyses are given in the supplemental material.

% ============================
% UPDATED Section 9: Discussion
% ============================
\rev{
\section{Discussions}
\label{sec:discussion}
}

\parag{\rev{Focus of this paper}}
\rev{
LLMs have been widely used in solving many problems and reasoning is an important ingredient in problem solving. As emerging models, the 
reasoning abilities and limitations of LLMs were unclear. 
Our main focus in this paper includes (i) a systematic empirical study of LLM behaviors on SAT and SAT-derived tasks and (ii) a new evaluation metric (ADR) for testing the faithful reasoning abilities of LLMs when they are applied to pairs of SAT formulas. }

\rev{
Our main insights include the following. First, our empirical study revealed that current LLMs have very limited reasoning ability as problem sizes grow bigger.
Second, ADR played an important role in uncovering these limitations.
In particular, under many settings LLM achieved high traditional metric values, but ADR revealed that such achievements were due to 
\SATLabel/\UNSAT\ skew, one-sided prediction bias, or reasoning based on superficial cues, such as clause density.
In this sense, ADR served a \emph{foundational evaluation signal}.
}

\parag{\rev{Opportunities for LLM--SAT Integration}}
\rev{
Our results revealed even reasoning LLMs work well only for small instances. Thus, an interesting way of leveraging LLMs is to
pursue \emph{solver centered integration}, where SAT solvers provide correctness and state management while LLMs are used only where they add measurable value. SAT solvers can help LLMs by (i) providing deterministic verification of decisions/witnesses and (ii) externalizing heavy symbolic operations (e.g., unit propagation, conflict checking, and state maintenance). This integration supports grounded and stepwise evaluation of LLMs and mitigates their scaling bottlenecks observed in our experiments.}

\rev{Conversely, LLMs may help SAT solvers as auxiliary policy components for solver control, such as branching heuristics, restart policies, or clause activity scoring. However, our results indicated that LLMs may provide seemingly correct suggestions while they are derived from 
superficial cues and can be misleading. Thus, before such integration efforts happen, the correctness of LLM suggestions should be carefully investigated. We envision that ADR can be an effective metric for such 
an investigation.
}

\parag{\rev{Applicability of ADR}}
\rev{
Applying ADR beyond synthetic SAT requires two conditions: (i) a deterministic checker for determining the labels of data instances and (ii) a principled perturbation tool that generates minimally different (or near minimal) pairs whose labels flip. 
% When they met, ADR can assess whether a model tracks logically critical edits rather than superficial cues. 
Thus, ADR applies naturally to many SAT derived settings such as graph problems through small edge edits, scheduling and configuration tasks through edits to hard constraints, and program analysis or verification through changes to specifications or source programs. }

\rev{As a proof of concept beyond random SAT formulas, we applied the same paired evaluation logic to SATLIB ``Flat'' graph colouring instances~\cite{hoos2000satlib,culberson2001frozen} (through graph level edits and SAT translation~\cite{karp2009reducibility}) and evaluated \texttt{GPT-5}. At $N{=}30$, conventional metrics were moderately high (Precision $=0.78$, Recall $=0.76$, F1 $=0.77$, Accuracy $=0.77$), but ADR was only $0.60$, again showing that single instance metrics can overstate reasoning abilities.}

\rev{ADR remains defined even for large instances.
Our results already show informative ADR trajectories at larger $N$ for some models (Figure 3). 
The main bottleneck is not the metric itself but the reliability and cost of obtaining outputs under limitations of prompt length, timeout, and API budget. In general, the problem of applying ADR where LLMs fail to respond reduces to the problem of how to scaling LLMs to deal with large instances. Thus, our idea outlined in leveraging SAT solvers for LLMs also apply here. An alternative approach is to measure the correctness of internal snapshots of an LLM' reasoning process (if supported by the LLM) rather than the final prediction result only.  
}

\parag{\rev{Limitations of ADR}}
\rev{
ADR is designed to measure a model's abilities 
to distinguish similar instances with different labels, which often indicates true reasoning abilities. Thus, a main limitation of ADR
is the availability of such paired instances. Fortunately, creating
paired instances can often be achieved by editing existing 
instances in small steps. A second limitation of ADR is that the labels of data instances must be accurate. This is, however, a general limitation with all metrics. A third limitation is that 
ADR is not designed to test 
end-to-end solver quality. In particular, ADR does not measure runtime, search efficiency, memory useage, or proof complexity.
It should also not be viewed as a replacement for solver centered performance metrics and should be viewed as a complement.
}

\section{Related Work}
\label{sec:rw}

\paragraph{Classical SAT, solvers, and phase transitions.}
The complexity-theoretic and algorithmic foundations of SAT are well established.
\textsc{2SAT} is solvable in linear time via implication graphs and SCCs~\citep{Aspvall:1979}, whereas \textsc{3SAT} is NP-complete~\citep{Cook:2023}.
Modern complete SAT solvers descend from DPLL~\citep{Davis:1962} and CDCL with clause learning and non-chronological backtracking~\citep{Marques:2002}, with subsequent engineering advances such as Chaff~\citep{Moskewicz:2001} and MiniSAT~\citep{Een:2003}.
The \emph{easy–hard–easy} phenomenon and the satisfiability phase transition at fixed clause density $\alpha=L/N$ have been documented for random $K$-SAT, especially $K{=}3$ near $\alpha_c\!\approx\!4.26$~\citep{Mitchell:1992,kirkpatrick1994critical}.
Our study uses these canonical behaviors as a ground truth for assessing whether LLMs exhibit solver-like sensitivity to instance structure.

\paragraph{LLMs for formal reasoning, specifications, and invariants.}
A growing body of work examines how far LLMs can go in program reasoning tasks.
\citet{CCSS:2024} show that raw LLM outputs are often logically weak in verification settings, limiting usefulness without additional structure.
For loop invariants, \citet{RLLI:2005} report that many attempts are typically needed to obtain a correct invariant and introduce a contrastive learning scheme to improve ranking; \citet{CLLM:2023} find that success rates correlate with the availability of small, informative traces.
\citet{Lemur:2023:Wu} integrate LLMs with automated verification backends and observe difficulties on complex formulas (e.g., nested conditionals and multi-loop cases), while \citet{EPSS:2024:Wen} show that unvalidated specification synthesis is frequently invalid and must be coupled with static analysis checks.
Overall, these papers point to substantial gaps between fluent generation and \emph{faithful} logical reasoning—an observation consistent with our findings on SAT.

\paragraph{LLMs, datasets, and evaluation pitfalls.}
Recent evaluations caution that headline metrics can be fragile under dataset noise, class imbalance, and prompt sensitivity.
In vulnerability detection, prior work reported strong numbers~(e.g., on popular benchmarks), but \citet{VDCL:Ding:2024} demonstrate that such gains can collapse on cleaner data: for instance, the F1 of StarCoder~2 drops drastically across datasets~\citep{SC2:Lozhkov:2024,VDCL:Ding:2024}.
Our results echo this methodological concern in the SAT setting: conventional accuracy/precision/recall/F1 can be inflated by the $\alpha$-dependent \SATLabel/\UNSAT\ skew, masking reasoning deficits.
This motivates our paired-design evaluation and ADR.
% , which are \need{robust to label imbalance.} \needAttention{It should be noted that ADR is not automatically robust to label imbalance across all datasets. Its robustness in our setting is a direct consequence of the paired evaluation design: 
By construction, each pair always contains one \SATLabel and one \UNSAT formula, thereby eliminating class-prior effects.

\paragraph{Reductions and cross-representation reasoning.}
Many real tasks reduce to SAT or SAT-like cores, including temporal network configuration~\citep{beckett2017general}, puzzles such as Sudoku~\citep{Yato:2003}, scheduling and planning~\citep{Even:1975,Kautz:1996}, and packing/placement problems~\citep{fekete2004combinatorial}.
Classical theory guarantees polynomial reductions among NP problems~\citep{Cook:2023}, but little empirical work has tested whether \emph{LLM decisions} remain stable under changes of representation (e.g., CNF $\leftrightarrow$ graphs/packing).
Our work fills this gap by measuring \emph{representation invariance} directly: we compare model decisions on CNF with their decisions on standard reductions to Vertex Cover and a discrete 3D-packing formulation, and we relate this to correctness.

Overall, compared to prior LLM-for-verification studies~\citep{CCSS:2024,RLLI:2005,CLLM:2023,Lemur:2023:Wu,EPSS:2024:Wen}, we focus on SAT as a \emph{foundational probe} with a well-understood difficulty landscape~\citep{Mitchell:1992,kirkpatrick1994critical} and mature solver baselines~\citep{Davis:1962,Marques:2002,Moskewicz:2001,Een:2003}.
Methodologically, we (i) highlight how standard metrics can be misleading under SAT/UNSAT imbalance, (ii) introduce a \emph{paired-formula} protocol with ADR to isolate structure-sensitive discrimination, and (iii) test \emph{cross-representation consistency} via canonical reductions (CNF $\rightarrow$ Vertex Cover, CNF $\rightarrow$ packing).
To our knowledge, this combination of phase-transition grounding, pairwise evaluation, and representation-invariance analysis has not been explored for LLM reasoning on SAT.

\section{Threats to Validity}
\label{sec:threats}

Potential threats include errors in reductions, parsing mistakes, or incorrect transformations into 3D-packing instances. We mitigated these by solver-verifying all CNFs, implementing unit-tested reductions, and enforcing strict JSON outputs to decouple reasoning from parsing errors. Traditional metrics (accuracy, precision, recall, F1, MCC) are biased by class imbalance or prediction asymmetry; instead, ADR directly measures pairwise correctness and is always defined. Model self-reported ``branches/conflicts'' were excluded because they did not align with solver depth.

External and statistical limitations include relatively small sample sizes and budget-constrained repetitions. However, by carefully designing our experiments, such as checking \UNSAT\ prediction and using paired formulas,
these instances are sufficient to capture large effects but may miss subtle differences. The study focused on random 2SAT/3SAT 
instances and a limited set of reductions (Vertex Cover, 3D packing), which may not fully represent real-world encodings. However, both problems are practical and are widely used in practice. Also, based on the theory of NP reductions, we believe that the results observed on them do transfer to 
other equally hard or harder problems.

\section{Conclusion}
\label{sec:con}

This paper examined whether large language models (LLMs) exhibit solver-like behavior on SAT and closely related formulations. Grounding our evaluation in phase-transition phenomena, we first showed that traditional metrics (accuracy/precision/recall/F1—and even MCC) are misleading under the $\alpha$-dependent \SATLabel/\UNSAT\ imbalance and strong model biases toward \SATLabel. LLM self-reported “branches/conflicts” did not track CDCL hardness, and performance degraded sharply with the number of variables~$N$, especially for \UNSAT\ detection.

To obtain a view of reasoning, we introduced \emph{paired formulas} and the \emph{Accurate Differentiation Rate} (ADR), which requires simultaneous correctness on minimally different \SATLabel/\UNSAT\ instances. ADR cleanly separated reasoning-oriented models from heuristic ones, declined smoothly with~$N$, and aligned with \emph{witness validity} (correct assignments/covers/packings). Beyond CNF, we tested representation invariance via reductions to \emph{Vertex Cover} and a discrete \emph{3D packing} formulation: decisions were largely consistent across representations, and the strongest model achieved both high consistency and truth alignment on small instances—yet all models remained scale-limited.

Current LLMs \emph{do not} solve SAT reliably at scale. Under paired evaluation, ADR is near~1 only for small $N$ and drops monotonically as $N$ grows; non-reasoning models typically fall below $\sim 0.2$ by $N\!\ge\!25$, and even the best reasoning-oriented models exhibit pronounced degradation. Thus, apparent gains under traditional metrics largely reflect distributional artifacts rather than durable, solver-like reasoning.

Overall, our results support SAT as a \emph{foundational} probe for LLM reasoning and advocate ADR with paired evaluation as a robust complement to conventional metrics. We view three directions as most promising: (i) scale-robust scaffolding that couples LLMs with verified search (e.g., CDCL-style tool use), (ii) training signals that reward pairwise discrimination and verifiable witnesses, and (iii) broader benchmarks that stress representation-invariant reasoning across standard reductions.

\begin{acks}
This material is based upon work supported by the National Science Foundation under Grant No.~1750886.
The authors gratefully acknowledge the Louisiana Optical Network Initiative (LONI) for providing the computational resources used in this study.
\end{acks}

\section*{Data-Availability Statement}
All datasets, prompts, supplemental material, and evaluation scripts used in this study are available in the archived Zenodo artifact~\cite{artifact:zenodo19524484}: 
\url{https://doi.org/10.5281/zenodo.19524484}. 
The corresponding development repository is available at:
\url{https://github.com/lzhan011/ADR_ESSL}.

\revdel{
\appendix
\section*{Appendix Overview}
}
\revdel{
This appendix provides additional technical details and supporting materials. 
Appendix develops the formal mathematical treatment of ADR, 
including complete proofs of the inequality chain 
($\max(0, r_S{+}r_U{-}1)\le \mathrm{ADR}\le \min(r_S,r_U)\le \mathrm{Acc}$), 
the accuracy identity $\mathrm{Acc}=\tfrac{1}{2}(r_S{+}r_U)$, 
and the independence decomposition 
$\mathrm{ADR}=r_Sr_U+\mathrm{Cov}(\mathbf{1}_A,\mathbf{1}_B)$. 
It also provides expanded counterexamples where accuracy/F1 appear artificially high 
while ADR correctly remains low.
}
\revdel{
Appendix presents the detailed derivations connecting ADR and 
the Matthews Correlation Coefficient (MCC). It shows the closed-form MCC in terms of recalls, 
$\mathrm{MCC}=\frac{r_S+r_U-1}{\sqrt{(r_S+1-r_U)(r_U+1-r_S)}}$, 
explains the undefined cases, and develops the pair-structured expression 
$\mathrm{MCC}=\frac{\mathrm{ADR}-\beta}{\sqrt{1-\delta^2}}$ with 
$\beta=\tfrac{n_{00}}{M}$ and $\delta=\tfrac{n_{10}-n_{01}}{M}$. 
Examples are included to illustrate how identical MCC values can mask 
substantial differences in ADR.
}
\revdel{
Appendix documents the CNF-to-Vertex Cover reduction used in our experiments. 
It provides the precise construction of graph vertices and edges from CNF clauses, 
together with the formal argument that the satisfiability of the original CNF 
corresponds exactly to the existence of a vertex cover of the required size. 
We also describe the deterministic checker used to validate candidate covers returned by models.
}
\revdel{
Finally, Appendix details the 3D packing experiments. 
It reproduces the exact one-line JSON prompt template used for packing feasibility queries 
and describes the deterministic witness validator that checks rod-selection 
and token-placement validity, ensuring both coverage and membership constraints are satisfied.
}

\bibliographystyle{ACM-Reference-Format}
\bibliography{paper}

@string{TOPLAS = "{ACM Trans.\ on Programming Languages and Systems}"}

@string{TOSEM  = "{ACM Trans.\ on Software Engineering and Methodology}"}

@article{Even:1975,
  title={On the complexity of time table and multi-commodity flow problems},
  author={Even, Shimon and Itai, Alon and Shamir, Adi},
  booktitle={16th annual symposium on foundations of computer science (sfcs 1975)},
  pages={184--193},
  year={1975},
  organization={IEEE}
}

@inproceedings{Kautz:1996,
  title={Pushing the envelope: Planning, propositional logic, and stochastic search},
  author={Kautz, Henry and Selman, Bart},
  booktitle={Proceedings of the national conference on artificial intelligence},
  pages={1194--1201},
  year={1996}
}

@article{Yato:2003,
  title={Complexity and completeness of finding another solution and its application to puzzles},
  author={Yato, Takayuki and Seta, Takahiro},
  journal={IEICE transactions on fundamentals of electronics, communications and computer sciences},
  volume={86},
  number={5},
  pages={1052--1060},
  year={2003},
  publisher={The Institute of Electronics, Information and Communication Engineers}
}

@inproceedings{Cook:2023,
  title={The complexity of theorem-proving procedures},
  author={Cook, Stephen A},
  booktitle={Logic, automata, and computational complexity: The works of Stephen A. Cook},
  pages={143--152},
  year={2023}
}

@inproceedings{CLLM:2023,
  title={Can large language models reason about program invariants?},
  author={Pei, Kexin and Bieber, David and Shi, Kensen and Sutton, Charles and Yin, Pengcheng},
  booktitle={International Conference on Machine Learning},
  pages={27496--27520},
  year={2023},
  organization={PMLR}
}

@article{RLLI:2005,
  title={Ranking llm-generated loop invariants for program verification},
  author={Chakraborty, Saikat and Lahiri, Shuvendu and Fakhoury, Sarah and Lal, Akash and Musuvathi, Madanlal and Rastogi, Aseem and Senthilnathan, Aditya and Sharma, Rahul and Swamy, Nikhil},
  booktitle={Findings of the Association for Computational Linguistics: EMNLP 2023},
  pages={9164--9175},
  year={2023}
}

@inproceedings{CCSS:2024,
  title={Can ChatGPT support software verification?},
  author={Jan{\ss}en, Christian and Richter, Cedric and Wehrheim, Heike},
  booktitle={International Conference on Fundamental Approaches to Software Engineering},
  pages={266--279},
  year={2024},
  organization={Springer}
}

@article{Lemur:2023:Wu,
  title={Lemur: Integrating large language models in automated program verification},
  author={Wu, Haoze and Barrett, Clark and Narodytska, Nina},
  journal={arXiv preprint arXiv:2310.04870},
  year={2023}
}

@inproceedings{EPSS:2024:Wen,
  title={Enchanting program specification synthesis by large language models using static analysis and program verification},
  author={Wen, Cheng and Cao, Jialun and Su, Jie and Xu, Zhiwu and Qin, Shengchao and He, Mengda and Li, Haokun and Cheung, Shing-Chi and Tian, Cong},
  booktitle={International Conference on Computer Aided Verification},
  pages={302--328},
  year={2024},
  organization={Springer},
  doi={10.1007/978-3-031-65630-9_16}
}

@inproceedings{LGIB:Pirzada:2024,
  title={Llm-generated invariants for bounded model checking without loop unrolling},
  author={Pirzada, Muhammad AA and Reger, Giles and Bhayat, Ahmed and Cordeiro, Lucas C},
  booktitle={Proceedings of the 39th IEEE/ACM International Conference on Automated Software Engineering},
  pages={1395--1407},
  year={2024},
  doi={10.1145/3691620.3695512}
}

@book{CPTT:2007:Alfred,
  title={Compilers: Principles, Techniques, and Tools},
  author={Aho, Alfred V. and Lam, Monica S. and Sethi, Ravi and Ullman, Jeffrey D.},
  year={2007},
  publisher={Pearson}
}

@article{IALL:2025:Raza,
  title={Industrial applications of large language models},
  author={Raza, Mubashar and Jahangir, Zarmina and Riaz, Muhammad Bilal and Saeed, Muhammad Jasim and Sattar, Muhammad Awais},
  journal={Scientific Reports},
  volume={15},
  number={1},
  pages={13755},
  year={2025},
  publisher={Nature Publishing Group UK London}
}

@misc{KACP:2025:Zheng,
  title={Knowledge augmented complex problem solving with large language models: A survey},
  author={Zheng, Da and Du, Lun and Su, Junwei and Tian, Yuchen and Zhu, Yuqi and Zhang, Jintian and Wei, Lanning and Zhang, Ningyu and Chen, Huajun},
  journal={arXiv preprint arXiv:2505.03418},
  year={2025}
}

@article{Wegman:1991:CPC,
  title={Constant propagation with conditional branches},
  author={Wegman, Mark N and Zadeck, F Kenneth},
  journal={ACM Transactions on Programming Languages and Systems (TOPLAS)},
  volume={13},
  number={2},
  pages={181--210},
  year={1991},
  publisher={ACM New York, NY, USA},
    doi={10.1145/103135.103136},
  url={https://doi.org/10.1145/103135.103136}
}

@incollection{BMC:2009:Biere,
  title={Bounded model checking.},
  author={Biere, Armin and Cimatti, Alessandro and Clarke, Edmund M and Strichman, Ofer and Zhu, Yunshan},
  journal={Handbook of satisfiability},
  volume={185},
  number={99},
  pages={457--481},
  year={2009}
}

@article{UA:1994:Ramalingam,
  title={The undecidability of aliasing},
  author={Ramalingam, Ganesan},
  journal={ACM Transactions on Programming Languages and Systems (TOPLAS)},
  volume={16},
  number={5},
  pages={1467--1471},
  year={1994},
  publisher={ACM New York, NY, USA},
  doi={10.1145/186025.186041}
}

@inproceedings{Reps:1995:PID,
  title={Precise interprocedural dataflow analysis via graph reachability},
  author={Reps, Thomas and Horwitz, Susan and Sagiv, Mooly},
  booktitle={Proceedings of the 22nd ACM SIGPLAN-SIGACT symposium on Principles of programming languages},
  pages={49--61},
  year={1995},
  doi={10.1145/199448.199462}
}

@article{CSL:Wang:2025,
  title={A contemporary survey of large language model assisted program analysis},
  author={Wang, Jiayimei and Ni, Tao and Lee, Wei-Bin and Zhao, Qingchuan},
  journal={arXiv preprint arXiv:2502.18474},
  year={2025}
}

@article{LLM:Hou:2024,
  title={Large language models for software engineering: A systematic literature review},
  author={Hou, Xinyi and Zhao, Yanjie and Liu, Yue and Yang, Zhou and Wang, Kailong and Li, Li and Luo, Xiapu and Lo, David and Grundy, John and Wang, Haoyu},
  journal={ACM Transactions on Software Engineering and Methodology},
  volume={33},
  number={8},
  pages={1--79},
  year={2024},
  publisher={ACM New York, NY},
  doi={10.1145/3695988}
}

@misc{SC2:Lozhkov:2024,
        title={Starcoder 2 and the stack v2: The next generation},
  author={Lozhkov, Anton and Li, Raymond and Allal, Loubna Ben and Cassano, Federico and Lamy-Poirier, Joel and Tazi, Nouamane and Tang, Ao and Pykhtar, Dmytro and Liu, Jiawei and Wei, Yuxiang and others},
  journal={arXiv preprint arXiv:2402.19173},
  year={2024}
}

@article{VDCL:Ding:2024,
  title={Vulnerability detection with code language models: How far are we?},
  author={Ding, Yangruibo and Fu, Yanjun and Ibrahim, Omniyyah and Sitawarin, Chawin and Chen, Xinyun and Alomair, Basel and Wagner, David and Ray, Baishakhi and Chen, Yizheng},
  journal={arXiv preprint arXiv:2403.18624},
  year={2024}
}

@article{Attention:Vaswani:2017,
  title={Attention is all you need},
  author={Vaswani, Ashish and Shazeer, Noam and Parmar, Niki and Uszkoreit, Jakob and Jones, Llion and Gomez, Aidan N and Kaiser, {\L}ukasz and Polosukhin, Illia},
  journal={Advances in neural information processing systems},
  volume={30},
  year={2017}
}

@article{ETA:Tay:2022,
  title={Efficient transformers: A survey},
  author={Tay, Yi and Dehghani, Mostafa and Bahri, Dara and Metzler, Donald},
  journal={ACM Computing Surveys},
  volume={55},
  number={6},
  pages={1--28},
  year={2022},
  publisher={ACM New York, NY}
}

@article{What:Clark:2019,
  title={What does BERT look at? an analysis of BERT’s attention},
  author={Clark, Kevin and Khandelwal, Urvashi and Levy, Omer and Manning, Christopher D},
  booktitle={Proceedings of the 2019 ACL workshop BlackboxNLP: analyzing and interpreting neural networks for NLP},
  pages={276--286},
  year={2019}
}

@article{Aspvall:1979,
  title={A linear-time algorithm for testing the truth of certain quantified boolean formulas},
  author={Aspvall, Bengt and Plass, Michael F and Tarjan, Robert Endre},
  journal={Information processing letters},
  volume={8},
  number={3},
  pages={121--123},
  year={1979},
  publisher={Elsevier}
}

@article{Davis:1962,
  title={A machine program for theorem-proving},
  author={Davis, Martin and Logemann, George and Loveland, Donald},
  journal={Communications of the ACM},
  volume={5},
  number={7},
  pages={394--397},
  year={1962},
  publisher={ACM New York, NY, USA},
  doi={10.1145/368273.368557}
}

@article{Marques:2002,
  title={GRASP: A search algorithm for propositional satisfiability},
  author={Marques-Silva, Joao P and Sakallah, Karem A},
  journal={IEEE Transactions on computers},
  volume={48},
  number={5},
  pages={506--521},
  year={2002},
  publisher={IEEE},
  doi={10.1109/12.769433}
}

@article{Tarjan:1972,
  title={Depth-first search and linear graph algorithms},
  author={Tarjan, Robert},
  journal={SIAM journal on computing},
  volume={1},
  number={2},
  pages={146--160},
  year={1972},
  publisher={SIAM}
}

@inproceedings{Moskewicz:2001,
  title={Chaff: Engineering an efficient SAT solver},
  author={Moskewicz, Matthew W and Madigan, Conor F and Zhao, Ying and Zhang, Lintao and Malik, Sharad},
  booktitle={Proceedings of the 38th annual Design Automation Conference},
  pages={530--535},
  year={2001}
}

@inproceedings{Een:2003,
  title={An extensible SAT-solver},
  author={E{\'e}n, Niklas and S{\"o}rensson, Niklas},
  booktitle={International conference on theory and applications of satisfiability testing},
  pages={502--518},
  year={2003},
  organization={Springer}
}

@inproceedings{Ball:2002:SLAM,
  title={The SLAM project: debugging system software via static analysis},
  author={Ball, Thomas and Rajamani, Sriram K},
  booktitle={Proceedings of the 29th ACM SIGPLAN-SIGACT symposium on Principles of programming languages},
  pages={1--3},
  year={2002},
  doi={10.1145/565816.503274}
}

@inproceedings{Henzinger:2002:LazyAbs,
  title={Lazy abstraction},
  author={Henzinger, Thomas A and Jhala, Ranjit and Majumdar, Rupak and Sutre, Gregoire},
  booktitle={Proceedings of the 29th ACM SIGPLAN-SIGACT symposium on Principles of programming languages},
  pages={58--70},
  year={2002}
}

@inproceedings{Engler:2001:Bugs,
  title={Bugs as deviant behavior: A general approach to inferring errors in systems code},
  author={Engler, Dawson and Chen, David Yu and Hallem, Seth and Chou, Andy and Chelf, Benjamin},
  journal={ACM SIGOPS Operating Systems Review},
  volume={35},
  number={5},
  pages={57--72},
  year={2001},
  publisher={ACM New York, NY, USA},
  doi={10.1145/502034.502041}
}

@inproceedings{shankar2001detecting,
  title={Detecting format string vulnerabilities with type qualifiers},
  author={Shankar, Umesh and Talwar, Kunal and Foster, Jeffrey S and Wagner, David},
  booktitle={10th USENIX Security Symposium (USENIX Security 01)},
  year={2001}
}

@inproceedings{Xie:2006:Buffer,
  title={Scalable error detection using boolean satisfiability},
  author={Xie, Yichen and Aiken, Alex},
  booktitle={Proceedings of the 32nd ACM SIGPLAN-SIGACT symposium on Principles of programming languages},
  pages={351--363},
  year={2005},
  doi={10.1145/1040305.1040334}
}

@article{Bessey:2010:Coverity,
  title={A few billion lines of code later: using static analysis to find bugs in the real world},
  author={Bessey, Al and Block, Ken and Chelf, Ben and Chou, Andy and Fulton, Bryan and Hallem, Seth and Henri-Gros, Charles and Kamsky, Asya and McPeak, Scott and Engler, Dawson},
  journal={Communications of the ACM},
  volume={53},
  number={2},
  pages={66--75},
  year={2010},
  publisher={ACM New York, NY, USA},
  doi={10.1145/1646353.1646374}
}

@inproceedings{Calcagno:2015:Infer,
  title={Moving fast with software verification},
  author={Calcagno, Cristiano and Distefano, Dino and Dubreil, J{\'e}r{\'e}my and Gabi, Dominik and Hooimeijer, Pieter and Luca, Martino and O’Hearn, Peter and Papakonstantinou, Irene and Purbrick, Jim and Rodriguez, Dulma},
  booktitle={NASA Formal Methods Symposium},
  pages={3--11},
  year={2015},
  organization={Springer}
}

@inproceedings{VulDeePecker:Li:2018,
  title={VulDeePecker: A deep learning-based system for vulnerability detection},
  author={Li, Zhen and Zou, Deqing and Xu, Shouhuai and Ou, Xinyu and Jin, Hai and Wang, Sujuan and Deng, Zhijun and Zhong, Yuyi},
  journal={arXiv preprint arXiv:1801.01681},
  year={2018}
}

@inproceedings{Devign:Zhou:2019,
  title={Devign: Effective vulnerability identification by learning comprehensive program semantics via graph neural networks},
  author={Zhou, Yaqin and Liu, Shangqing and Siow, Jingkai and Du, Xiaoning and Liu, Yang},
  journal={Advances in neural information processing systems},
  volume={32},
  year={2019}
}

@article{zou2019mu,
  title={{$\mu$}VulDeePecker: A Deep Learning-Based System for Multiclass Vulnerability Detection},
  author={Zou, Deqing and Wang, Sujuan and Xu, Shouhuai and Li, Zhen and Jin, Hai},
  journal={IEEE Transactions on Dependable and Secure Computing},
  volume={18},
  number={5},
  pages={2224--2236},
  year={2019},
  publisher={IEEE},
  doi={10.1109/TDSC.2019.2942930}
}

@inproceedings{Mitchell:1992,
  title={Hard and easy distributions of SAT problems},
  author={Mitchell, David and Selman, Bart and Levesque, Hector and others},
  booktitle={AAAI},
  volume={92},
  pages={459--465},
  year={1992}
}

@article{DeepWukong:Cheng:2021,
  title={Deepwukong: Statically detecting software vulnerabilities using deep graph neural network},
  author={Cheng, Xiao and Wang, Haoyu and Hua, Jiayi and Xu, Guoai and Sui, Yulei},
  journal={ACM Transactions on Software Engineering and Methodology (TOSEM)},
  volume={30},
  number={3},
  pages={1--33},
  year={2021},
  publisher={ACM New York, NY, USA},
  doi={10.1145/3436877}
}

@article{powers2011evaluation,
  title={Evaluation: from precision, recall and F-measure to ROC, informedness, markedness and correlation},
  author={Powers, David MW},
  journal={arXiv preprint arXiv:2010.16061},
  year={2020}
}

@book{manning2008ir,
  title={Introduction to information retrieval},
  author={Manning, Christopher D},
  year={2008},
  publisher={Syngress Publishing}
}

@article{kirkpatrick1994critical,
  title={Critical behavior in the satisfiability of random boolean expressions},
  author={Kirkpatrick, Scott and Selman, Bart},
  journal={Science},
  volume={264},
  number={5163},
  pages={1297--1301},
  year={1994},
  publisher={American Association for the Advancement of Science}
}

@article{matthews1975mcc,
  title={Comparison of the predicted and observed secondary structure of T4 phage lysozyme},
  author={Matthews, Brian W},
  journal={Biochimica et Biophysica Acta (BBA)-Protein Structure},
  volume={405},
  number={2},
  pages={442--451},
  year={1975},
  publisher={Elsevier}
}

@article{chicco2020mcc,
  title={The advantages of the Matthews correlation coefficient (MCC) over F1 score and accuracy in binary classification evaluation},
  author={Chicco, Davide and Jurman, Giuseppe},
  journal={BMC genomics},
  volume={21},
  number={1},
  pages={6},
  year={2020},
  publisher={Springer}
}

@article{hoos2000satlib,
  title={SATLIB: An online resource for research on SAT},
  author={Hoos, Holger H and St{\"u}tzle, Thomas},
  journal={Sat},
  volume={2000},
  pages={283--292},
  year={2000}
}

@article{culberson2001frozen,
  title={Frozen development in graph coloring},
  author={Culberson, Joseph and Gent, Ian},
  journal={Theoretical computer science},
  volume={265},
  number={1-2},
  pages={227--264},
  year={2001},
  publisher={Elsevier}
}

@incollection{karp2009reducibility,
  title={Reducibility among combinatorial problems},
  author={Karp, Richard M},
  booktitle={50 Years of Integer Programming 1958-2008: from the Early Years to the State-of-the-Art},
  pages={219--241},
  year={2009},
  publisher={Springer}
}

@article{fekete2004combinatorial,
  title={A combinatorial characterization of higher-dimensional orthogonal packing},
  author={Fekete, S{\'a}ndor P and Schepers, J{\"o}rg},
  journal={Mathematics of Operations Research},
  volume={29},
  number={2},
  pages={353--368},
  year={2004},
  publisher={INFORMS}
}

@inproceedings{beckett2017general,
  title={A general approach to network configuration verification},
  author={Beckett, Ryan and Gupta, Aarti and Mahajan, Ratul and Walker, David},
  booktitle={Proceedings of the Conference of the ACM Special Interest Group on Data Communication},
  pages={155--168},
  year={2017},
  doi={10.1145/3098822.3098834}
}

@misc{artifact:zenodo19524484,
  author       = {Zhang, Leizhen and Chen, Shuhan and Chen, Sheng},
  title        = {{Evaluating Satisfiability Solving with LLMs}},
  year         = {2026},
  publisher    = {Zenodo},
  doi          = {10.5281/zenodo.19524484},
  url          = {https://doi.org/10.5281/zenodo.19524484}
}


\end{document}